\documentclass[]{bytedance}
\usepackage[toc,page,header]{appendix}

\usepackage{minitoc}
\usepackage{amsfonts}
\usepackage{amssymb}
\usepackage{tabularx}
\usepackage{listings}
\usepackage{xcolor}
\usepackage{cancel}

\usepackage{tabulary,multirow,xspace}
\usepackage{fixmath,mathtools,nicefrac,mmstyle}
\usepackage{subcaption}
\usepackage{caption}
\usepackage{float}
\usepackage{wrapfig} %
\usepackage[misc]{ifsym} %
\usepackage{colortbl}

\usepackage{multicol}
\usepackage[most]{tcolorbox}
\usepackage{pifont}
\usepackage{amsmath}
\usepackage{amssymb}
\usepackage{graphicx}
\usepackage{makecell}
\usepackage{todonotes}
\usepackage{enumitem}
\usepackage{threeparttable}
\usepackage[table]{xcolor}
\usepackage{booktabs}
\usepackage{tabularx}
\usepackage{multirow}
\usepackage{subcaption}
\usepackage{wrapfig}
\usepackage{comment}
\usepackage{array}
\usepackage{rotating}
\usepackage{adjustbox}
\usepackage{listings}
\usepackage{graphicx}
\usepackage{subcaption}

\definecolor{codegreen}{rgb}{0,0.6,0}
\definecolor{codegray}{rgb}{0.5,0.5,0.5}
\definecolor{codepurple}{rgb}{0.58,0,0.82}
\definecolor{backcolour}{rgb}{0.95,0.95,0.92}
\definecolor{boxblue}{RGB}{57,89,163}
\definecolor{boxbluebg}{RGB}{230,237,250} 

\lstdefinestyle{mystyle}{
    backgroundcolor=\color{backcolour},   
    commentstyle=\color{codegreen},
    keywordstyle=\color{magenta},
    numberstyle=\tiny\color{codegray},
    stringstyle=\color{codepurple},
    basicstyle=\ttfamily\footnotesize,
    breakatwhitespace=false,         
    breaklines=true,                 
    captionpos=b,                    
    keepspaces=true,                 
    numbers=none,                    
    numbersep=5pt,                  
    showspaces=false,                
    showstringspaces=false,
    showtabs=false,                  
    tabsize=2
}
\definecolor{mygray1}{gray}{.95}
\definecolor{mygray2}{gray}{.9}
\definecolor{mygray3}{gray}{.95}
\usepackage{pifont}%

\newlength\savewidth
\newcolumntype{x}[1]{>{\centering\arraybackslash}p{#1pt}}

\newcommand{\app}{\raise.17ex\hbox{$\scriptstyle\sim$}}

\usepackage{xcolor}
\usepackage{listings}

\definecolor{commentgreen}{rgb}{0.1, 0.4, 0.1}
\definecolor{keywordblue}{rgb}{0.1, 0.1, 0.7}
\definecolor{stringred}{rgb}{0.7, 0.1, 0.1}

\lstdefinestyle{mystyle}{
    commentstyle=\color{commentgreen},
    keywordstyle=\color{keywordblue},   
    stringstyle=\color{stringred},
    basicstyle=\ttfamily\scriptsize, 
    breaklines=true,
    keepspaces=true,
    showstringspaces=false,
    frame=none,                     
    language=Python, 
}

\newcommand{\name}{SparseGDN}
\title{\name{}: Chunk Sparse in Linear Attention}

\author[1,2,*]{Jiedong Zhuang}
\author[2]{Peng Zhao}
\author[2]{Jialin Tang}
\author[1,\dagger]{Haoji Hu}

\affiliation[1]{Zhejiang University}
\affiliation[2]{ByteDance}

\contribution[*]{Work done at ByteDance}
\contribution[\dagger]{Corresponding authors}

\label{sec:abstract}
\abstract{
Gated Delta Networks (GDNs) offer linear-time sequence modeling, yet
long-context prefill remains costly due to dense intra-chunk computation,
recurrent-state updates, and intermediate state storage.
We introduce \textsc{SparseGDN}, a hardware-aware framework that combines
structured intra-chunk approximation with input-dependent recurrent-event
selection while producing outputs for every token.
A block-adjacent solver reduces local Delta-solve computation, while a
dynamic selector allocates recurrent-update budgets across head pairs.
The GPU implementation materializes updates only for selected events and
reconstructs token outputs using compact state snapshots and group-local
interactions, reducing computation and memory traffic.
We evaluate \textsc{SparseGDN} on Qwen3.5-4B and Qwen3.5-9B across
RULER tasks and context lengths from 4K to 128K.
Retaining approximately \(\textbf{29}\%\) of recurrent events on average,
\textsc{SparseGDN} achieves \(\textbf{1.50}\)--\(\textbf{1.94}\times\) GDN-core prefill
speedups over native FLA kernels 
, including
online selection overhead.
}

\date{\today} \correspondence{\email{haoji\_hu@zju.edu.cn}}

\begin{document}
\maketitle

\begin{figure*}[ht]
  \centering
  \begin{subfigure}[t]{0.49\textwidth}
      \centering
      \includegraphics[width=\linewidth]{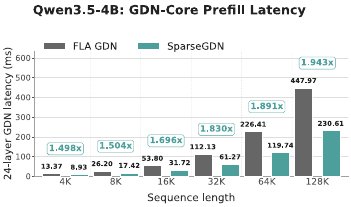}
      \label{fig:qwen4b}
  \end{subfigure}
  \hfill
  \begin{subfigure}[t]{0.49\textwidth}
      \centering
      \includegraphics[width=\linewidth]{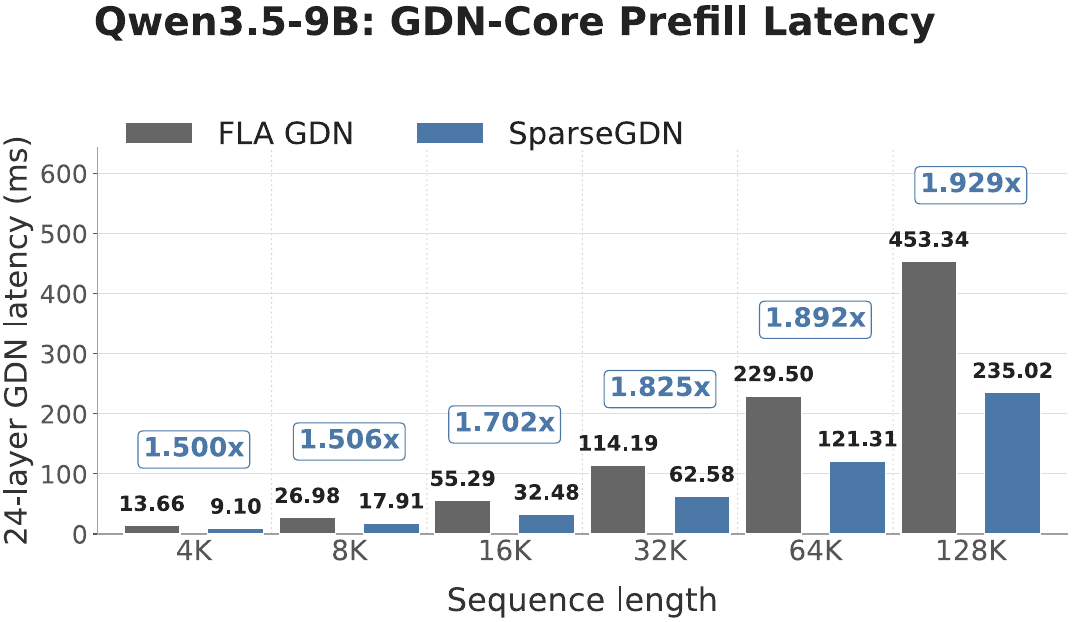}
      \label{fig:qwen9b}
  \end{subfigure}

  \caption{
    GDN-core prefill latency across context lengths for Qwen3.5-9B and Qwen3.5-4B.
    Compared with FLA GDN, \textsc{SparseGDN} additionally includes the online selector overhead. Latency is aggregated over all 24 GDN layers. \textsc{SparseGDN} reduces latency at every context length, reaching \(1.93\times\) and \(1.94\times\) speedup at 128K for the 9B and 4B models, respectively. Lower is better.}
  \label{fig:speedup_qwen}
\end{figure*}

\section{Introduction}
\label{sec:introduction}

The quadratic cost of self-attention remains a major obstacle to efficient
long-context prefill in Transformer-based language
models~\citep{vaswani2017attention,dao2022flashattention,
jiang2024minference,yuan2025nsa}.
Recurrent and state-space architectures offer an alternative by summarizing
historical information in a fixed-size state and processing sequences with
linear complexity~\citep{katharopoulos2020transformers,gu2022efficiently,
gu2024mamba,dao2024transformers}.
Gated Delta Networks (GDNs) combine input-dependent state updates with
expressive associative-memory dynamics~\citep{schlag2021linear,
yang2024deltarule,yang2024gateddeltanet}.
Nevertheless, linear asymptotic complexity alone does not guarantee
efficient execution.
During long-context prefill, dense GDN implementations perform local
Delta solves within every chunk, materialize update matrices, propagate
recurrent states across chunks and heads, and reconstruct token outputs
from dense state snapshots.
These operations incur substantial arithmetic and memory-traffic costs
at context lengths of tens or hundreds of thousands of tokens.

\begin{figure*}[t]
  \centering
  \begin{subfigure}[t]{0.55\textwidth}
      \centering
      \includegraphics[width=\linewidth]{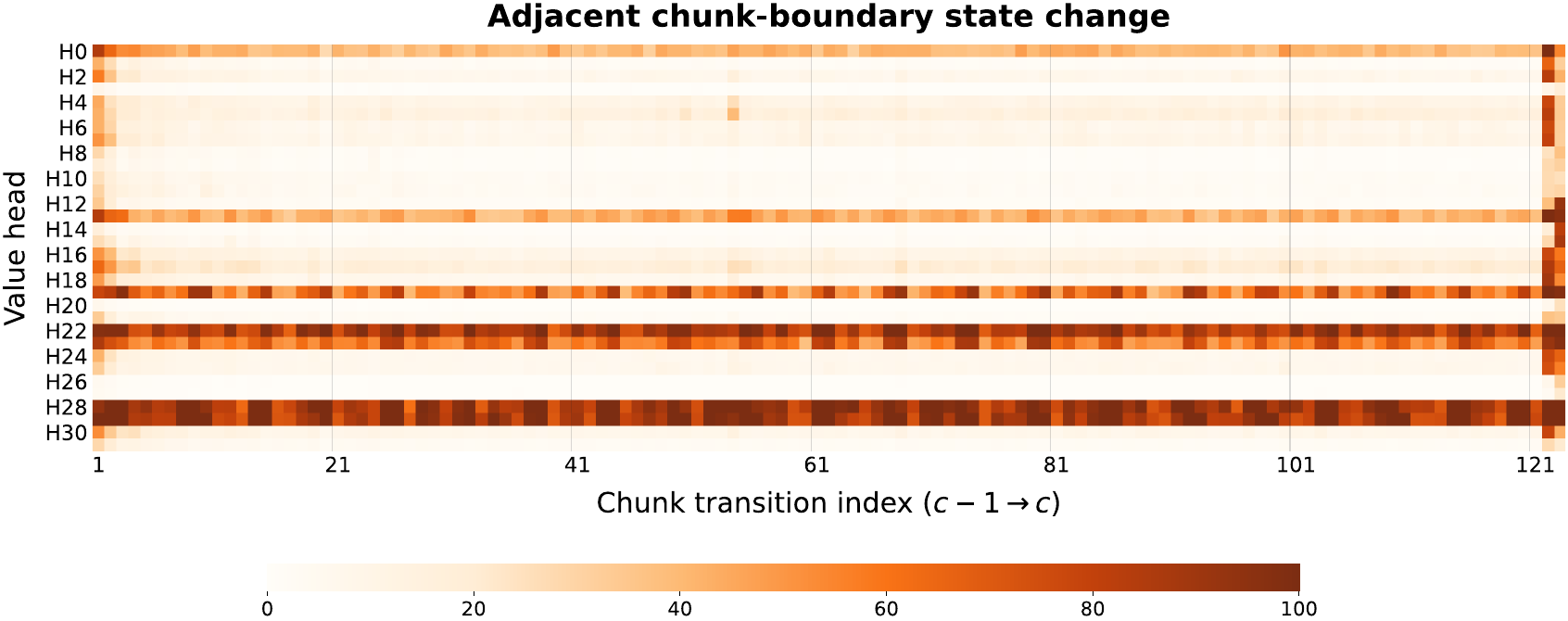}
      \caption{Inter-chunk recurrent-state changes across value heads.}
      \label{fig:state_change}
  \end{subfigure}
  \hfill
  \begin{subfigure}[t]{0.43\textwidth}
      \centering
      \includegraphics[width=\linewidth]{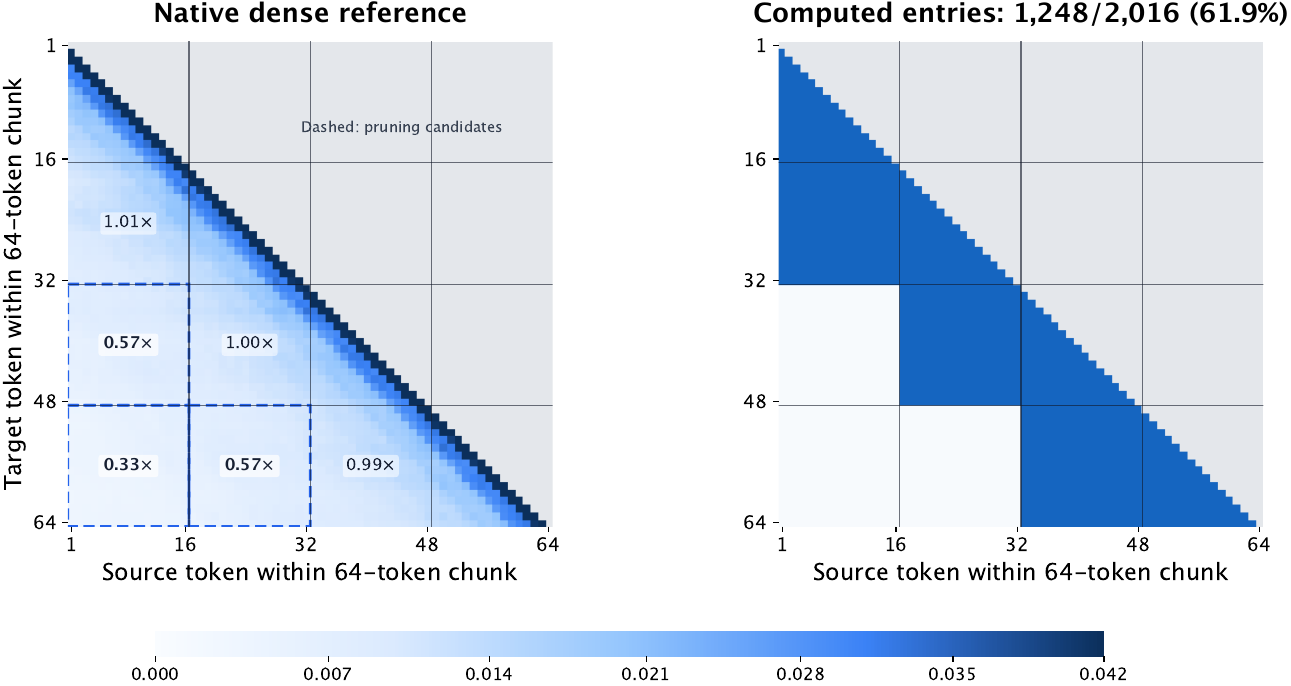}
      \caption{Dense intra-chunk coefficients and block-adjacent support.}
      \label{fig:intra_structure}
  \end{subfigure}

  \caption{
      Two empirical structures observed in layer 16 of native-dense
      Qwen3.5-4B on an 8K RULER input
      (\texttt{niah\_multikey\_3}, row 0).
      \textbf{(a)} Adjacent chunk-boundary states are frequently nearly
      stationary, while the magnitudes of their changes vary substantially
      across value heads.
      \textbf{(b)} The left panel shows
      $\lvert\mathbf{M}^{-1}-\mathbf{I}\rvert$, averaged over
      123 complete 64-token chunks and all 32 value heads.
      Here $\mathbf{M}$ is the native Delta-solve matrix defined in
      Section~\ref{sec:blockadj}; subtracting $\mathbf{I}$ removes the unit
      diagonal, and absolute values are taken elementwise before averaging.
      The coefficients concentrate near the causal diagonal and weaken
      across more distant token blocks.
      This observed locality motivates the block-adjacent approximation,
      whose computation support is shown in the right panel.
  }
  \label{fig:sparsegdn_motivation}
\end{figure*}

Figure~\ref{fig:sparsegdn_motivation} illustrates two empirical structures
that motivate sparsification in GDN prefill.
At the \emph{inter-chunk} level, we examine the recurrent state
$S_{c,h}$ of value head $h$ at chunk boundary $c$ and measure its
symmetric relative change:
\[
t_{c,h}
=
\frac{\lVert S_{c,h}-S_{c-1,h}\rVert_F}
{\max\!\left(
  \lVert S_{c,h}\rVert_F,
  \lVert S_{c-1,h}\rVert_F
\right)}.
\]
For the layer and input shown in Figure~\ref{fig:state_change},
the median state change is $6.56\%$,
$63.8\%$ of head--chunk transitions change by less than $10\%$,
and the median cosine similarity between adjacent states is $0.9979$.
These state dynamics vary substantially across heads:
21 of 32 value heads have a median change below $10\%$, whereas a
smaller subset exhibits large and persistent changes.
Related head heterogeneity has also motivated distinct cache policies for
retrieval and streaming attention heads~\citep{xiao2025duoattention}.
Small state changes alone do not establish that an update can be removed
safely, but the observed variation motivates testing retention policies
that adapt to different inputs, layers, and heads.

At the \emph{intra-chunk} level,
Figure~\ref{fig:intra_structure} summarizes the mean absolute
off-diagonal coefficients of the native 64-token Delta solve across
chunks and value heads.
The coefficients are strongest near the causal diagonal
and become weaker across more distant 16-token blocks.
This locality motivates restricting the support of the local inverse
to diagonal and immediately adjacent blocks.

Building on these observations, we introduce \textsc{SparseGDN},
a two-level sparsification framework for GDN prefill.
A block-adjacent solver reduces intra-chunk Delta-solve computation,
while a dynamic selector allocates retention budgets across head pairs
and selects recurrent events.
Selection uses tail activation statistics, recurrent lifetime estimates,
and output-gate information available before recurrent propagation,
avoiding the need to execute candidate updates before deciding
which to retain.

To realize GPU speedups from this sparsity~\citep{gale2023megablocks,
zhang2025spargeattention,ye2025flashinfer},
\textsc{SparseGDN} materializes update matrices only for selected events
and packs them into a compressed sparse row (CSR) layout.
The recurrent kernel scans retained events, aggregates decay across
skipped chunks, and stores compact state snapshots.
A mapped output kernel combines these snapshots with group-local
causal interactions to produce outputs for every token.

We evaluate \textsc{SparseGDN} on Qwen3.5-4B and Qwen3.5-9B using the
RULER long-context benchmark~\citep{hsieh2024ruler}.
Our evaluation covers 13 tasks and six context lengths from 4K to 128K.
Across these settings, \textsc{SparseGDN} retains approximately
$29\%$ of recurrent pair--chunk events on average.
It achieves
$1.50\times$--$1.94\times$ GDN-core prefill speedups for Qwen3.5-4B
and $1.50\times$--$1.93\times$ for Qwen3.5-9B.
These measurements include online selection overhead and aggregate latency
over all 24 GDN layers in each model.
The gains increase with context length.

Our main contributions are:

\begin{itemize}
  \item We identify complementary intra-chunk coefficient locality and
  inter-chunk state heterogeneity, and develop a two-level sparsification
  strategy that combines a block-adjacent Delta solve with input-dependent
  recurrent-event selection.

  \item We design a GPU execution pipeline with deferred update
  materialization, CSR-based event scanning, compact state storage,
  and mapped output reconstruction to reduce computation and memory
  traffic while producing outputs for every token.

  \item We demonstrate consistent GDN-core prefill acceleration on
  Qwen3.5-4B and Qwen3.5-9B over context lengths from 4K to 128K,
  while maintaining comparable aggregate performance across RULER tasks.
\end{itemize}

\section{Method}
\label{sec:method}

We present \textsc{SparseGDN}, a two-level sparsification framework for
long-context Gated Delta Network prefill. The method combines
(i) structured block sparsity within each chunk and
(ii) dynamically selected recurrent events across chunks.
We first describe the dense GDN formulation and then introduce each stage of
the sparse execution pipeline.

\begin{figure}[t]
    \centering
    \includegraphics[width=\linewidth]{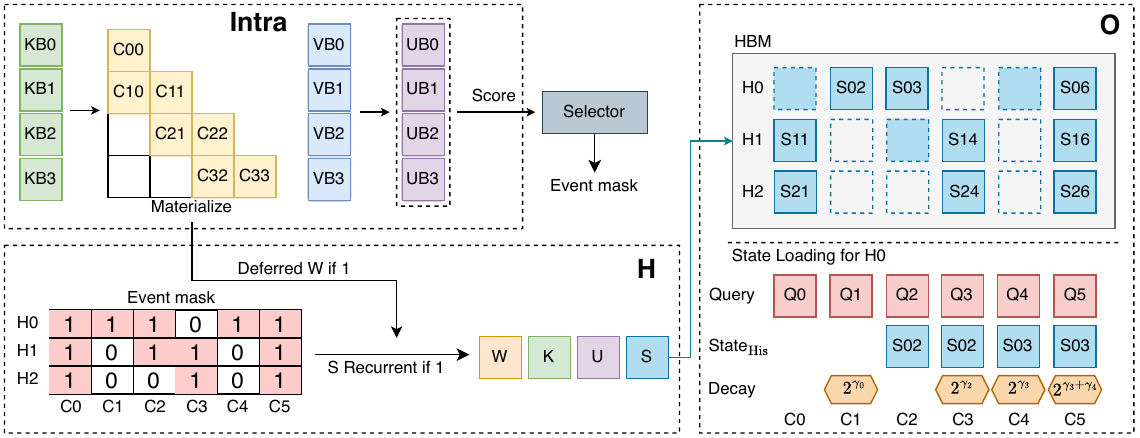}
\caption{\textbf{End-to-end SparseGDN execution pipeline.}
  The block-adjacent intra-chunk kernel computes transformed values $\mathbf{U}$, compact update coefficients, and
  selector scores. The dynamic selector converts these scores into a structured event mask. Only selected events trigger
  deferred $\mathbf{W}$ materialization and recurrent-state updates, producing compact event states. During output
  reconstruction, the mapped output kernel retrieves the appropriate preceding event state from HBM and applies the
  accumulated gap decay, while computing group-local interactions and producing outputs
  for every token.}
\label{fig:intra+H+O}
\end{figure}

\subsection{Dense GDN Prefill}
\label{sec:dense_gdn}

For clarity, we omit the batch dimension. Let the query, key, and value
tensors be
\[
\mathbf{Q},\mathbf{K}\in\mathbb{R}^{T\times N_q\times d},
\qquad
\mathbf{V}\in\mathbb{R}^{T\times N_v\times d_v},
\]
where \(T\) is the sequence length. In Qwen3.5-4B and Qwen3.5-9B,
\(N_q=16\), \(N_v=32\), and \(d=d_v=128\). Each query/key head is shared
by two adjacent value heads. For a fixed value head \(h\), we write
\(\mathbf{Q}_c\) and \(\mathbf{K}_c\) for its associated query/key matrices,
suppressing their shared head index.

The input sequence is partitioned into chunks of \(L=64\) tokens.
For chunk \(c\) and value head \(h\), the dense intra-chunk operator
produces a transformed value matrix
\(\mathbf{U}_{c,h}\in\mathbb{R}^{L\times d_v}\)
and a read-correction matrix
\(\mathbf{W}_{c,h}\in\mathbb{R}^{L\times d}\).
Both matrices are obtained from the local Delta solve,
whose formulation is given in Section~\ref{sec:blockadj}.
Let \(\mathbf{S}_{c,h}\in\mathbb{R}^{d\times d_v}\) denote the recurrent
state before processing chunk \(c\). The state-corrected values are
\begin{equation}
\mathbf{U}'_{c,h}
=
\mathbf{U}_{c,h}
-
\mathbf{W}_{c,h}\mathbf{S}_{c,h}.
\label{eq:dense_value_correction}
\end{equation}

Let \(\mathbf{g}_{c,h}\in\mathbb{R}^{L}\) be the chunk-local cumulative gate
in log-base-two space, and let
\(\gamma_{c,h}=g_{c,h,L-1}\) be its final value.
Thus, \(2^{g_{c,h,i}}\) is the cumulative retention from the start of
chunk \(c\) through token \(i\), and
\(2^{\gamma_{c,h}-g_{c,h,i}}\) is the retention from just after token \(i\)
to the chunk boundary.
Exponentials act elementwise; in products with matrices, \(\odot\)
broadcasts each gate vector across feature channels to scale token rows.
The recurrent state update is
\begin{equation}
\mathbf{S}_{c+1,h}
=
2^{\gamma_{c,h}}\mathbf{S}_{c,h}
+
\mathbf{K}_{c}^{\top}
\left(
  \mathbf{U}'_{c,h}
  \odot
  2^{\gamma_{c,h}-\mathbf{g}_{c,h}}
\right).
\label{eq:dense_state_update}
\end{equation}

The GDN-core output \(\mathbf{O}_{c,h}\) precedes the model's RMSNorm,
output gating, and output projection. It combines the historical recurrent
state with causal local interactions:
\begin{equation}
\mathbf{O}_{c,h}
=
\frac{1}{\sqrt{d}}
\left[
  2^{\mathbf{g}_{c,h}}
  \odot
  \left(\mathbf{Q}_{c}\mathbf{S}_{c,h}\right)
  +
  \mathbf{A}_{c,h}\mathbf{U}'_{c,h}
\right],
\label{eq:dense_output}
\end{equation}
where
\begin{equation}
\mathbf{A}_{c,h}[i,j]
=
\left\langle
  \mathbf{Q}_{c}[i],
  \mathbf{K}_{c}[j]
\right\rangle
2^{g_{c,h,i}-g_{c,h,j}}
\mathbb{I}[j\leq i].
\label{eq:local_attention}
\end{equation}

An incomplete final chunk is padded with zero keys, values, and update
strengths and unit per-token retention, so padded positions neither write
to nor decay the state. Only outputs for the original \(T\) tokens are
returned.

Although the computation scales linearly with \(T\), the dense chunkwise
implementation materializes \(\mathbf{W}_{c,h}\), performs recurrent
updates, and stores intermediate states for every chunk and value head.

\subsection{Method Overview and Approximation Boundary}
\label{sec:method_overview}

\textsc{SparseGDN} sparsifies GDN prefill at two levels.
Within each chunk, it replaces the full Delta solve with a
block-adjacent approximation.
Across chunks, it applies recurrent read corrections and state writes
only at dynamically selected events.
Each recurrent event corresponds to one chunk and one pair of adjacent
value heads, which share an event mask.
State decay is preserved across skipped chunks, and outputs are
computed for every token.

Output reconstruction operates on consecutive two-chunk groups
\((2r,2r+1)\).
Within each group, the odd chunk's output retains the preceding even
chunk's local contribution even when that even chunk is skipped by
the recurrent stage.
This local contribution does not enter the persistent state or the
subsequent recurrent read correction.
The group-local output rule is therefore part of the sparse operator
definition, as detailed in Section~\ref{sec:mapped_o} and
Equation~\eqref{eq:group2_local_read_difference}.

For the resulting sparse operator, deferred materialization of
\(\mathbf{W}\), CSR-based recurrent propagation, compact state storage,
and mapped output evaluation are execution optimizations.
They preserve the operator's algebraic definition while reducing
intermediate materialization and state traffic.

The full pipeline consists of
\[
\text{Block-Adjacent Intra}
\rightarrow
\text{Dynamic Selector}
\rightarrow
\text{Deferred-}W
\rightarrow
\text{Event-Scan }H
\rightarrow
\text{Mapped }O.
\]

\subsection{Block-Adjacent Intra-Chunk Computation}
\label{sec:blockadj}

For a fixed chunk and value head, we omit their indices.
Let \(\boldsymbol{\beta}\in\mathbb{R}^{L}\) contain the token-wise
update strengths. The native local Delta solve uses
\[
\begin{aligned}
\mathbf{M}&=\mathbf{I}+\mathbf{A}_{\Delta},\\
\mathbf{A}_{\Delta}[t,s]
&=\beta_t\langle\mathbf{K}[t],\mathbf{K}[s]\rangle
  2^{g_t-g_s}\mathbb{I}[s<t],
\qquad 0\leq t,s<L.
\end{aligned}
\]
The dense transformed values and read-correction matrix solve
\[
\mathbf{M}\mathbf{U}
=\operatorname{diag}(\boldsymbol{\beta})\mathbf{V},
\qquad
\mathbf{M}\mathbf{W}
=\operatorname{diag}(\boldsymbol{\beta}\odot2^{\mathbf{g}})\mathbf{K}.
\]
The update strengths therefore enter both the interaction matrix and
the right-hand sides.

Motivated by the locality observed in Figure~\ref{fig:intra_structure},
we divide each 64-token chunk into four consecutive blocks of \(b=16\)
tokens. We approximate \(\mathbf{M}^{-1}\) by a block matrix
\(\mathbf{C}\) supported on four diagonal blocks and three immediately
adjacent cross-blocks:
\[
\mathcal{B}
=
\{(0,0),(1,1),(2,2),(3,3),(1,0),(2,1),(3,2)\}.
\]
Let \(\mathbf{M}_{ij}\) denote the corresponding \(16\times16\) block
of \(\mathbf{M}\). For each diagonal block, we compute
\begin{equation}
\mathbf{C}_{ii}
=
\mathbf{M}_{ii}^{-1},
\qquad i\in\{0,1,2,3\},
\label{eq:blockadj_diagonal}
\end{equation}
and for each adjacent block pair,
\begin{equation}
\mathbf{C}_{i,i-1}
=
-\mathbf{C}_{ii}
\mathbf{M}_{i,i-1}
\mathbf{C}_{i-1,i-1},
\qquad i\in\{1,2,3\}.
\label{eq:blockadj_offdiagonal}
\end{equation}
All other blocks of \(\mathbf{C}\) are zero.
In exact arithmetic, the retained blocks coincide with the corresponding
blocks of \(\mathbf{M}^{-1}\); the approximation omits the more distant
lower-triangular blocks.
This support retains 1,248 of the 2,016 causal off-diagonal entries
in a 64-token chunk.

From this point onward, \(\mathbf{U}\) and \(\mathbf{W}\) denote the
block-adjacent approximations to their dense counterparts.
The seven retained blocks are reused to form two sets of weighted
coefficients. For \((i,j)\in\mathcal{B}\), define
\[
\begin{aligned}
\widetilde{\mathbf{C}}_{ij}
&=\mathbf{C}_{ij}\operatorname{diag}(\boldsymbol{\beta}_j),\\
\overline{\mathbf{C}}_{ij}
&=\widetilde{\mathbf{C}}_{ij}\operatorname{diag}(2^{\mathbf{g}_j})
 =\mathbf{C}_{ij}
  \operatorname{diag}(\boldsymbol{\beta}_j\odot2^{\mathbf{g}_j}).
\end{aligned}
\]
Here \(\boldsymbol{\beta}_j\) and \(\mathbf{g}_j\) are the restrictions
of the chunk's update-strength and cumulative-gate vectors to source
block \(j\). Right multiplication scales the columns of
\(\mathbf{C}_{ij}\).
Thus, \(\mathbf{C}\) contains inverse-solve coefficients,
\(\widetilde{\mathbf{C}}\) generates \(\mathbf{U}\), and
\(\overline{\mathbf{C}}\) generates \(\mathbf{W}\).

Let \(\mathbf{UB}_i\), \(\mathbf{VB}_i\), \(\mathbf{KB}_i\), and
\(\mathbf{WB}_i\) denote the \(i\)-th 16-token blocks of
\(\mathbf{U}\), \(\mathbf{V}\), \(\mathbf{K}\), and \(\mathbf{W}\),
respectively, following the block labels in
Figure~\ref{fig:intra+H+O}.
The block-adjacent transformed values are
\begin{align}
\mathbf{UB}_0
&=
\widetilde{\mathbf{C}}_{00}\mathbf{VB}_0,
\\
\mathbf{UB}_i
&=
\widetilde{\mathbf{C}}_{ii}\mathbf{VB}_i
+
\widetilde{\mathbf{C}}_{i,i-1}\mathbf{VB}_{i-1},
\qquad i\in\{1,2,3\}.
\label{eq:blockadj_u}
\end{align}
The four transformed-value blocks are computed and stored progressively,
reducing the number of simultaneously live accumulator tiles.

The Intra kernel stores the seven compact blocks of
\(\overline{\mathbf{C}}\) for deferred \(\mathbf{W}\) materialization,
while consuming \(\widetilde{\mathbf{C}}\) locally to compute
\(\mathbf{U}\).
It also computes a tail statistic while \(\mathbf{UB}_3\) remains
on chip. Restoring the chunk and head indices, this statistic is
\begin{equation}
s_{c,h}
=
\sqrt{
  \sum_{t=48}^{63}
  \sum_{v=0}^{d_v-1}
  U_{c,h,t,v}^{2}
}.
\label{eq:tail_score}
\end{equation}

Tail-16 uses transformed-value magnitude near the chunk boundary as
a low-cost proxy for recurrent-write strength.
In Equation~\eqref{eq:dense_state_update}, each token's contribution
is weighted by \(2^{\gamma_{c,h}-g_{c,h,t}}\).
Because per-token retentions lie in \((0,1]\), later tokens undergo
no more cumulative attenuation than earlier ones.
The values \(\mathbf{U}\) already incorporate update strengths and
the retained local Delta interactions, allowing scoring before
\(\mathbf{W}\) materialization and the state-dependent correction
\(\mathbf{W}\mathbf{S}\).
The 16-token window matches the final tile \(\mathbf{UB}_3\), so its
Frobenius norm can be computed without rereading \(\mathbf{U}\)
from HBM.

This score is a ranking heuristic.
The actual recurrent write also depends on the keys, decay weights,
and corrected values
\(\mathbf{U}'=\mathbf{U}-\mathbf{W}\mathbf{S}\).
Tail magnitude alone does not bound the write magnitude or downstream
error, and important contributions may occur earlier in the chunk.
We therefore use it as the base signal in a selector whose budget
allocation also accounts for recurrent lifetime and output visibility
(Section~\ref{sec:selector}).
Section~\ref{sec:algorithmic_ablations} evaluates its diagnostic ranking
quality, and the left panel of Figure~\ref{fig:tail_window_ablation}
compares ranking recall and standalone score-kernel cost across window
sizes.
The choice of 16 tokens is an empirical and implementation trade-off.

\begin{figure}[t]
    \centering
    \includegraphics[width=\linewidth]{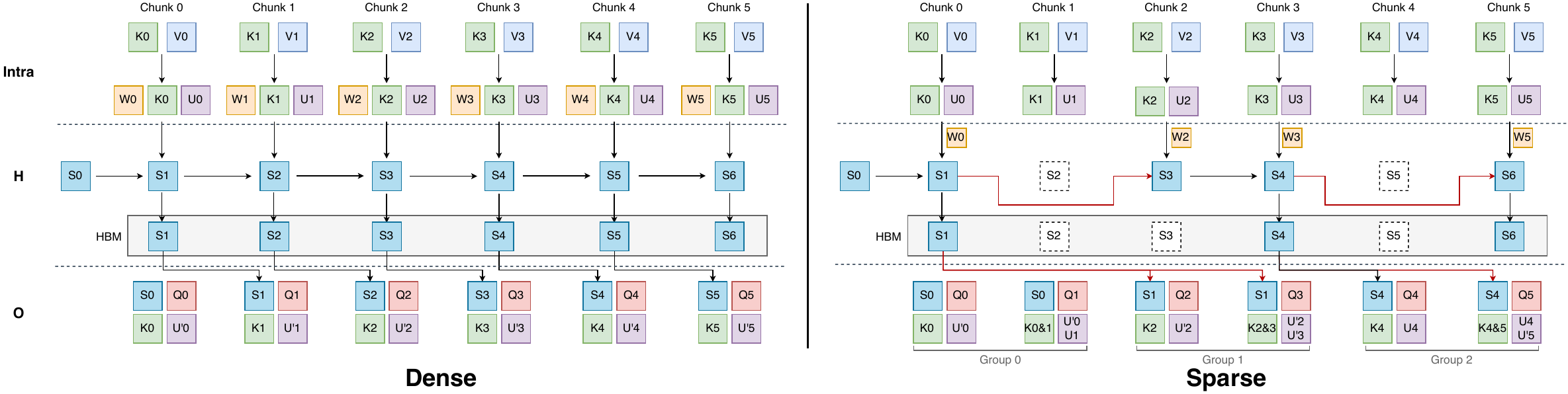}
    \caption{\textbf{Dense versus SparseGDN execution.}
    The dense chunkwise implementation materializes \(\mathbf{W}\),
    updates the recurrent state, and stores a chunk-boundary state
    for every chunk.
    SparseGDN materializes \(\mathbf{W}\) and performs recurrent
    read/write only at selected events; accumulated decay bridges
    skipped chunks (red arrows), and only intermediate state snapshots
    required by downstream output groups are stored in HBM.
    The final recurrent state is returned for decoding.
    Mapped group-2 reconstruction reuses one group-start state for two
    adjacent chunks while still producing outputs for every token.
    Dashed boxes denote skipped computation or elided state storage.}
    \label{fig:denseVSsparse}
\end{figure}

\subsection{Dynamic Event Selection}
\label{sec:selector}

\begin{figure}[t]
    \centering
    \includegraphics[width=\linewidth]{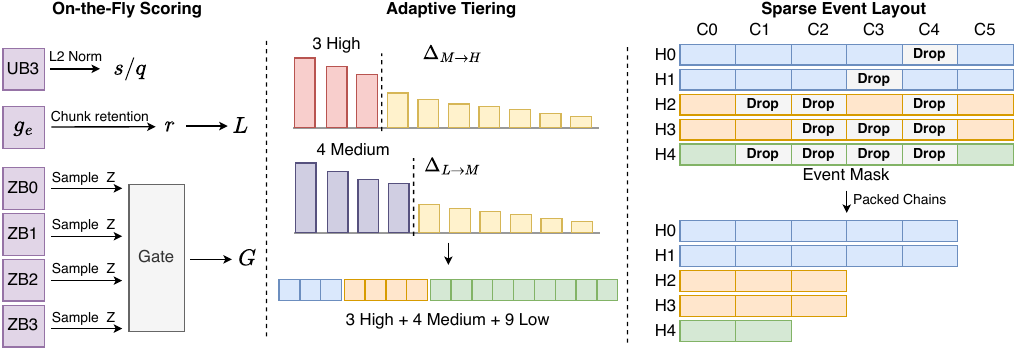}
    \caption{\textbf{Dynamic event selection and packing.}
    Here, \(\mathbf{ZB}_i\) denotes the \(i\)-th 16-token block of the
    pre-activation output-gate tensor \(Z\) within a chunk.
    Chunk scores determine the retention order within each head pair.
    Recurrent lifetime and output-visibility proxies weight the estimated
    gains from larger budgets, which determine the low-, medium-, and
    high-retention tiers.
    The example assigns 9, 4, and 3 pairs to these tiers.
    Selected events are packed in temporal order into CSR chains
    for recurrent execution.}
    \label{fig:selector}
\end{figure}

The selector determines which recurrent events to retain and how much
computation each head pair receives.
It first ranks chunks within each pair, estimates the benefit of larger
budgets, and then assigns each pair a low-, medium-, or high-retention tier
(Figure~\ref{fig:selector}).
Let \(C=C_{\mathrm{valid}}=\lceil T/64\rceil\) be the number of valid chunks.
There are 16 head pairs per layer.
Only valid chunks participate in selection; workspace padding is excluded.
When \(C=1\), the sole chunk is retained.
The following formulas assume \(C\geq2\).
For clarity, numerical safeguards are omitted; normalization maps
all-zero vectors to zero.

\subsubsection{Rank chunks within each head pair}

For the ranking, upgrade-gain, and mask formulas, we fix one head pair
and omit its index.
Sums over \(h\) in these formulas run only over its two value heads.
We normalize the Intra score \(s_{c,h}\) from
Equation~\eqref{eq:tail_score} separately for each head and average:
\begin{equation}
q_c
=
\frac{1}{2}\sum_h
\frac{s_{c,h}}{\|s_{\cdot,h}\|_2}.
\label{eq:pair_rank_score}
\end{equation}
This per-head normalization balances the two heads' contributions
to chunk ranking.

Each pair has three candidate budgets, with retention rates
\[
(\rho_{\mathrm L},\rho_{\mathrm M},\rho_{\mathrm H})
=
(0.08,0.30,0.95).
\]
Tier \(\ell\) keeps
\[
k_\ell
=
\operatorname{clip}
\bigl(\operatorname{round}(\rho_\ell C),2,C\bigr)
\]
chunks.
The first and last chunks are mandatory anchors, and the remaining
\(k_\ell-2\) chunks are selected by \(q_c\).
Using the same ranking gives nested candidate sets
\[
\mathcal C_{\mathrm L}
\subseteq
\mathcal C_{\mathrm M}
\subseteq
\mathcal C_{\mathrm H}.
\]

\subsubsection{Estimate the benefit of higher retention}

We weight the additional scores retained by a larger budget using
head-wise estimates of recurrent persistence and output visibility.
Let \(r_h\) be the mean chunk-end retention \(2^{\gamma_{c,h}}\)
over the last \(\min(16,C)\) valid chunks.
Using \(r_h\) as a constant per-chunk retention, we construct a
truncated geometric persistence weight with a common horizon of
\(C-2\) chunks.

To estimate output visibility, we sample entries of the pre-activation
output-gate tensor \(Z\).
This tensor is projected from the layer input and is available before
recurrent propagation; it subsequently modulates the normalized
GDN-core output before output projection.
Let \(\eta_h\) be the mean squared magnitude of the sampled entries
after SiLU and multiplication by the learned RMSNorm scale.
The lifetime and relative visibility weights are
\begin{equation}
L_h
=
\sum_{j=1}^{C-2}r_h^j,
\qquad
G_h
=
\sqrt{\frac{\eta_h}{\bar{\eta}}},
\qquad
\bar{\eta}
=
\frac{1}{N_v}\sum_{j=0}^{N_v-1}\eta_j.
\label{eq:head_lifetime}
\end{equation}
Here, \(\bar{\eta}\) averages gate energy over all value heads in the
same layer and request.
The lifetime weight estimates persistence due to gate decay under
the constant-retention approximation; it does not model how subsequent
Delta updates modify existing information.
The visibility weight measures relative gate energy:
\(G_h=1\) at average energy, with larger values indicating stronger
gate scaling.

For either upgrade
\(a\!\to\!b\in
\{\mathrm L\!\to\!\mathrm M,\mathrm M\!\to\!\mathrm H\}\),
we sum the weighted scores of the additional chunks it would retain:
\begin{equation}
\Delta_{a\rightarrow b}
=
\sum_{c\in\mathcal C_b\setminus\mathcal C_a}
\sum_h s_{c,h}L_hG_h.
\label{eq:tier_gain}
\end{equation}
Unlike the normalized scores used for chunk ranking, this gain uses
the original \(s_{c,h}\) magnitudes to preserve strength differences
across heads and pairs.
It is a heuristic proxy for the benefit of increasing a pair's budget.

Repeating this calculation for all pairs, we normalize each upgrade-gain
vector by its mean across pairs, obtaining \(x_{\mathrm M}\) for
\(\mathrm L\!\to\!\mathrm M\) and \(x_{\mathrm H}\) for
\(\mathrm M\!\to\!\mathrm H\).
Together, these two vectors form a \(16\times2\) array of normalized
upgrade gains.

\subsubsection{Assign head-pair budgets}

We assign high-retention pairs first, then medium-retention pairs
from those remaining.
Both decisions use the same split criterion.
For a descending vector \(x\) of \(n\) gains, define
\begin{equation}
B(x,k)
=
\frac{k(n-k)}{n}
\left(
\overline{x}_{1:k}
-
\overline{x}_{k+1:n}
\right)^2,
\qquad 1\leq k<n,
\label{eq:high_cluster_variance}
\end{equation}
where each bar denotes the mean over the indicated ranks.
This criterion measures the separation between the higher- and
lower-gain groups induced by a cut.

First, sort \(x_{\mathrm H}\) and choose
\(n_{\mathrm H}\in\{1,2,3,4\}\) to maximize
\(B(x_{\mathrm H},n_{\mathrm H})\).
The first \(n_{\mathrm H}\) pairs receive the high tier.
Then sort \(x_{\mathrm M}\) over the remaining
\(16-n_{\mathrm H}\) pairs and apply the same criterion within
the following candidate counts:
\[
\begin{array}{c|cccc}
n_{\mathrm H} & 1 & 2 & 3 & 4\\
\hline
n_{\mathrm M} & \{11,12,13\} & \{7,8,9\} & \{3,4,5\} & \{0\}
\end{array}
\]
For \(n_{\mathrm H}=4\), set \(n_{\mathrm M}=0\) directly.
Assign the first \(n_{\mathrm M}\) remaining pairs to medium
and all others to low.

The nominal retention rate is
\[
\rho_{\mathrm{nominal}}
=
\frac{
n_{\mathrm H}\rho_{\mathrm H}
+
n_{\mathrm M}\rho_{\mathrm M}
+
(16-n_{\mathrm H}-n_{\mathrm M})\rho_{\mathrm L}
}{16}.
\]
The candidate middle-tier counts are centered on
\(4(4-n_{\mathrm H})\), keeping nominal retention near \(30\%\).
Allowing neighboring counts gives a nominal range of
\(28.44\%\)--\(31.31\%\).
The selected tier counts, integer rounding, and anchors determine
the actual keep rate, which averages approximately \(29\%\)
in our experiments.

For the fixed pair, let \(\ell\) be its assigned tier.
Its event mask is
\begin{equation}
z_c
=
\mathbb I[c\in\mathcal C_\ell].
\label{eq:final_event_mask}
\end{equation}
Both value heads share this mask while maintaining independent
recurrent states.

\paragraph{Gate sampling.}
Gate energy samples one token position per 16-token block,
at zero-based offsets \(8,24,40,56\) within each valid chunk.
At each sampled position, it reads one rotating value coordinate
of \(Z\) per head:
\[
\eta_h
=
\frac{1}{4C}
\sum_{c=0}^{C-1}
\sum_{a=0}^{3}
\left[
\nu_{v_{c,a}}
\operatorname{SiLU}
\left(Z_{t_{c,a},h,v_{c,a}}\right)
\right]^2,
\]
where \(\nu\) is the learned RMSNorm scale,
\(t_{c,a}=64c+8+16a\), and
\(v_{c,a}=73(4c+a)\bmod d_v\).
Out-of-range samples are zero.

\paragraph{Numerical and boundary conventions.}
The geometric sum for \(L_h\) approaches \(C-2\) as \(r_h\)
approaches one.
Equal ranking scores prefer smaller chunk or pair indices,
and keep counts use ties-to-even rounding.
Numerically tied split scores prefer the count closest to the
fallback, then the smaller count.
Flat scores use fallback counts of 3 for high and
\(4(4-n_{\mathrm H})\) for medium.
The workspace pads to \(2\lceil C/2\rceil\) chunks without
adding selection candidates.

\subsection{Deferred Update Materialization and Event Layout}
\label{sec:deferred_w}

The dense chunkwise implementation materializes
\(\mathbf{W}_{c,h}\) for every chunk and value head.
\textsc{SparseGDN} stores the seven weighted coefficient blocks
\(\overline{\mathbf{C}}\) during Intra
(Section~\ref{sec:blockadj}) and materializes \(\mathbf{W}\)
after selection, only for retained events.

For a selected chunk and value head, the four 16-token blocks of
\(\mathbf{W}\) are
\begin{align}
\mathbf{WB}_0
&=
\overline{\mathbf{C}}_{00}\mathbf{KB}_0,
\\
\mathbf{WB}_i
&=
\overline{\mathbf{C}}_{ii}\mathbf{KB}_i
+
\overline{\mathbf{C}}_{i,i-1}\mathbf{KB}_{i-1},
\qquad i\in\{1,2,3\}.
\label{eq:deferred_w}
\end{align}
Here, \(\mathbf{KB}_j\) is the source key block.
Since
\(\overline{\mathbf{C}}_{ij}
=\mathbf{C}_{ij}\operatorname{diag}
(\boldsymbol{\beta}_j\odot2^{\mathbf{g}_j})\)
already incorporates the source-token update strengths and cumulative
gates, this stage multiplies the stored coefficients by the
corresponding key blocks.

The selected events are organized in a compressed sparse row (CSR)
layout, with one row per head pair.
Rows are ordered by tier from high to medium to low, with increasing
physical head-pair indices within each tier.
Within each row, retained chunks appear in increasing temporal order.
This arrangement prioritizes longer event chains during recurrent
execution.
The event representation and its output-mapping metadata contain:
\begin{itemize}
  \item \texttt{pair\_order}, which maps each CSR row to its physical
  head pair;
  \item \texttt{event\_offsets}, which delimit the event interval
  for each CSR row;
  \item \texttt{event\_indices}, which store retained chunk indices
  in temporal order within each row;
  \item \texttt{group\_event\_map}, which maps each head pair and output
  group to the packed index of the latest retained event strictly
  before the group starts, using \(-1\) when no such event exists; and
  \item \texttt{decay\_prefix}, which stores per-value-head prefix sums
  of chunk-end log-retentions for gap-decay reconstruction.
\end{itemize}

The number of selected pair--chunk events is
\[
E
=
\sum_{\ell\in\{\mathrm L,\mathrm M,\mathrm H\}}
n_\ell k_\ell,
\]
where \(n_\ell\) is the number of head pairs assigned to tier \(\ell\),
and \(k_\ell\) is the corresponding number of retained chunks per pair.

We distinguish this valid event count from the allocated capacity.
The host reserves space for the case in which every pair receives
the high tier:
\[
E_{\max}
=
\frac{N_v}{2}k_{\mathrm H}
=
16k_{\mathrm H},
\qquad
\mathbf{W}^{\mathrm{packed}}
\in
\mathbb{R}^{E_{\max}\times64\times2\times d}.
\]
Only the first \(E\) event slots contain valid read-correction matrices,
which are written directly in CSR event order.
The two value heads are interleaved within each token row.
The final CSR offset stores the actual count \(E\) on the GPU,
allowing subsequent kernels to identify valid event intervals without
copying \(E\) to the host.
Thus, buffer allocation does not require a device-to-host
synchronization to determine the selected event count.

\begin{figure}[t]
    \centering
    \includegraphics[width=\linewidth]{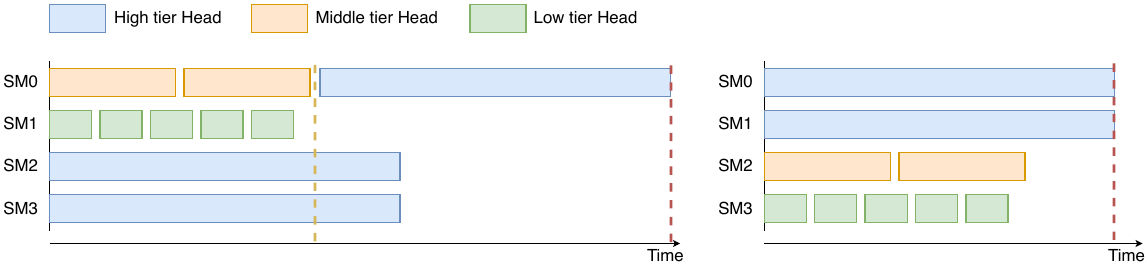}
    \caption{\textbf{Tier-aware scheduling of heterogeneous recurrent event chains.} High-, medium-, and low-retention head pairs induce event chains with different workloads. Scheduling them in physical-head order can place a long high-tier chain behind shorter tasks, creating an SM straggler and extending kernel completion time (left). SparseGDN orders chains by expected workload and uses event-count-adaptive execution to distribute them across SMs (right), reducing load imbalance and tail latency. The red dashed line denotes kernel completion.}
\label{fig:sm}
\end{figure}

\subsection{Event-Sparse Recurrent Propagation}
\label{sec:event_h}

For zero-initialized prefill, \(\mathbf{S}_{0,h}=\mathbf{0}\).
Using the block-adjacent Intra outputs, the recurrent stage processes
the selected events for each head pair in temporal order.
For a retained event at chunk \(c\), each value head \(h\) applies
the recurrent read correction and state write:
\begin{align}
\mathbf{U}'_{c,h}
&=
\mathbf{U}_{c,h}
-
\mathbf{W}_{c,h}\mathbf{S}_{c,h},
\\
\mathbf{S}_{c+1,h}
&=
2^{\gamma_{c,h}}\mathbf{S}_{c,h}
+
\mathbf{K}_{c}^{\top}
\left(
  \mathbf{U}'_{c,h}
  \odot
  2^{\gamma_{c,h}-\mathbf{g}_{c,h}}
\right).
\label{eq:sparse_retained_update}
\end{align}
For a skipped event, the sparse recurrence is defined by
\begin{equation}
\mathbf{U}'_{c,h}=\mathbf{U}_{c,h},
\qquad
\mathbf{S}_{c+1,h}
=
2^{\gamma_{c,h}}\mathbf{S}_{c,h}.
\label{eq:sparse_dropped_update}
\end{equation}
Thus, skipping an event preserves state decay while omitting the
read correction and the write of new content into the persistent state.
Token outputs are still produced for the chunk by the output stage.

The kernel avoids visiting skipped chunks individually by using
prefix sums of chunk-end log-retentions:
\begin{equation}
P_{0,h}=0,
\qquad
P_{c+1,h}
=
\sum_{j=0}^{c}\gamma_{j,h}.
\label{eq:decay_prefix}
\end{equation}
These prefixes include both retained and skipped chunks.
For consecutive retained events at chunks \(e<c\), the decay across
the intervening skipped chunks is
\begin{equation}
D_{\mathrm{gap}}(e,c,h)
=
2^{P_{c,h}-P_{e+1,h}}
=
\prod_{j=e+1}^{c-1}2^{\gamma_{j,h}}.
\label{eq:gap_decay}
\end{equation}
The state entering chunk \(c\) is therefore
\[
\mathbf{S}_{c,h}
=
D_{\mathrm{gap}}(e,c,h)\mathbf{S}_{e+1,h}.
\]
For adjacent retained chunks, the empty product equals one.
Decay within chunk \(c\) is applied by
Equation~\eqref{eq:sparse_retained_update}.
The recurrent kernel scans each CSR event chain and applies one
accumulated decay per gap, preserving the algebraic recurrence defined
by Equation~\eqref{eq:sparse_dropped_update}.

The transformed-value buffer is reused in place.
It initially stores \(\mathbf{U}\), and the recurrent kernel overwrites
only retained events with their corrected values \(\mathbf{U}'\).
Skipped events remain unchanged and already satisfy
\(\mathbf{U}'=\mathbf{U}\).
The same buffer therefore supplies the corrected values for all chunks
to the output stage.

The scan carries the current recurrent state between events.
Among intermediate states, only post-update snapshots
\(\mathbf{S}_{c+1,h}\) referenced by later output groups are stored
in the compact event-state buffer.
The final state \(\mathbf{S}_{C,h}\) is returned separately to
initialize autoregressive decoding.
This representation reduces intermediate state stores while preserving
the snapshots needed for mapped output reconstruction
(Section~\ref{sec:mapped_o}).

\subsection{Mapped Group-Wise Output Reconstruction}
\label{sec:mapped_o}

\textsc{SparseGDN} produces outputs for every token by combining states
from sparse recurrent propagation with group-local interactions.
We fix one value head and omit its index.
Group \(r\) consists of chunks \(c=2r\) and \(c+1\), and
\(\mathbf{S}_c\) denotes the state before chunk \(c\).
The mask \(z_c\) is shared by the corresponding value-head pair.
The output stage reads the \(\mathbf{U}'\) buffer from
Section~\ref{sec:event_h}, with \(\mathbf{U}'_c=\mathbf{U}_c\)
for skipped events.
If the number of valid chunks is odd, the final group is completed
with a neutral padding chunk under the convention in
Section~\ref{sec:dense_gdn}.
Only outputs for the original \(T\) tokens are returned.

For the current head pair and output group,
\texttt{group\_event\_map} locates the latest retained event before
the group starts.
If that event occurs at chunk \(e<c\), its stored post-update snapshot
is \(\mathbf{S}_{e+1}\).
The group-entry state is recovered using the decay prefix:
\[
\mathbf{S}_c
=
2^{P_c-P_{e+1}}\mathbf{S}_{e+1}.
\]
With zero-initialized prefill, a group with no preceding event uses
\(\mathbf{S}_c=\mathbf{0}\).

For the local interaction matrices below, the two subscripts identify
the output chunk and source chunk, respectively.
Let \(\mathbf{A}_{c,c}\) denote the within-chunk causal matrix from
Equation~\eqref{eq:local_attention}.
The adjacent cross-chunk matrix is
\[
\mathbf{A}_{c+1,c}[i,j]
=
\left\langle\mathbf{Q}_{c+1}[i],\mathbf{K}_c[j]\right\rangle
2^{g_{c+1,i}+\gamma_c-g_{c,j}},
\qquad 0\leq i,j<L.
\]
All tokens in chunk \(c\) precede those in chunk \(c+1\), so this
cross-chunk matrix requires no additional triangular mask.

Both outputs reuse the group-entry state:
\begin{equation}
\mathbf{O}_c
=
\frac{1}{\sqrt{d}}
\left[
2^{\mathbf{g}_c}\odot(\mathbf{Q}_c\mathbf{S}_c)
+
\mathbf{A}_{c,c}\mathbf{U}'_c
\right],
\label{eq:group2_even}
\end{equation}
\begin{equation}
\begin{aligned}
\mathbf{O}_{c+1}
=
\frac{1}{\sqrt{d}}\bigl[
&2^{\mathbf{g}_{c+1}+\gamma_c}
\odot(\mathbf{Q}_{c+1}\mathbf{S}_c)\\
&+\mathbf{A}_{c+1,c}\mathbf{U}'_c
+\mathbf{A}_{c+1,c+1}\mathbf{U}'_{c+1}
\bigr].
\end{aligned}
\label{eq:group2_odd_expansion}
\end{equation}
The odd-chunk historical term includes the decay across chunk \(c\).
Its two local terms use the concatenated keys and corrected values
of the group.
The adjacent term is included independently of \(z_c\).
Direct local interactions remain within the group; earlier history
is represented by \(\mathbf{S}_c\).

The persistent state and recurrent read correction still follow
Section~\ref{sec:event_h}:
\begin{equation}
\begin{aligned}
\mathbf{S}_{c+1}
&=
2^{\gamma_c}\mathbf{S}_c
+
z_c\mathbf{K}_c^{\top}
\left(
\mathbf{U}'_c\odot2^{\gamma_c-\mathbf{g}_c}
\right),
\\
\mathbf{U}'_{c+1}
&=
\begin{cases}
\mathbf{U}_{c+1}-\mathbf{W}_{c+1}\mathbf{S}_{c+1},
& z_{c+1}=1,\\
\mathbf{U}_{c+1},
& z_{c+1}=0.
\end{cases}
\end{aligned}
\label{eq:group2_sparse_state}
\end{equation}
Thus, the local contribution from a skipped even chunk affects
the group output without entering this layer's persistent state
or the odd chunk's recurrent read correction.

To isolate the effect of the output rule, fix the Intra outputs
and event mask.
Let \(\mathbf{O}^{\mathrm{chunk}}_{c+1}\) denote
Equation~\eqref{eq:dense_output} evaluated using the same sparse state
\(\mathbf{S}_{c+1}\) and corrected values \(\mathbf{U}'_{c+1}\).
Subtracting this per-chunk output from the group output gives
\begin{equation}
\mathbf{O}_{c+1}-\mathbf{O}^{\mathrm{chunk}}_{c+1}
=
\frac{1-z_c}{\sqrt{d}}\,
\mathbf{A}_{c+1,c}\mathbf{U}'_c.
\label{eq:group2_local_read_difference}
\end{equation}
When \(z_c=1\), the two output rules agree algebraically for the
same sparse state and corrected values.
When \(z_c=0\), the group output additionally retains
\(\mathbf{A}_{c+1,c}\mathbf{U}_c/\sqrt{d}\), since
\(\mathbf{U}'_c=\mathbf{U}_c\).
This group-local read is part of the sparse operator defined in
Section~\ref{sec:method_overview}.

For example, Figure~\ref{fig:denseVSsparse} skips chunk 4 and retains
chunk 5.
The persistent state and corrected values satisfy
\[
\mathbf{S}_5=2^{\gamma_4}\mathbf{S}_4,
\qquad
\mathbf{U}'_5=\mathbf{U}_5-\mathbf{W}_5\mathbf{S}_5.
\]
The local inputs for \(\mathbf{Q}_5\) are
\([\mathbf{K}_4;\mathbf{K}_5]\) and
\([\mathbf{U}_4;\mathbf{U}'_5]\), where the semicolon denotes
concatenation along the token dimension.
The output therefore includes
\(\mathbf{A}_{5,4}\mathbf{U}_4/\sqrt{d}\), while the recurrent
correction uses the decay-only state \(\mathbf{S}_5\).

The kernel fuses the adjacent and within-chunk local terms and reuses
the reconstructed group-entry state for both outputs.
This avoids materializing a separate intermediate state snapshot
for the odd chunk's output.

\section{Experiments}
\label{sec:experiments}

We evaluate \textsc{SparseGDN} along two dimensions: long-context task
performance and GDN-core inference efficiency. Experiments are conducted on
both Qwen3.5-4B and Qwen3.5-9B to evaluate whether the proposed sparsification
strategy generalizes across model scales.

\subsection{Experimental Setup}
\label{sec:experimental_setup}

\paragraph{Models.}
We evaluate Qwen3.5-4B and Qwen3.5-9B. Both models contain
24 GDN layers with the same recurrent geometry: 16 query/key heads,
32 value heads, and key and value head dimensions of 128.
Model activations and tensor-core matrix multiplication inputs use BF16,
while recurrent-state accumulation and selector statistics use FP32.

\paragraph{Dataset.}
We use the RULER long-context benchmark~\citep{hsieh2024ruler},
which evaluates retrieval, multi-hop tracing, aggregation, and
question answering.
We evaluate all 13 tasks at six context lengths:
4K, 8K, 16K, 32K, 64K, and 128K.
Each task--length configuration contains 100 examples,
giving 1,300 examples per length and 7,800 examples per model.



\paragraph{Efficiency measurement.}
We measure GDN-core prefill latency 
using PyTorch 2.9.0+cu128, Triton 3.5.0, and Transformers 5.9.0.
Unless stated otherwise, the batch size is one.
These measurements use a fixed timing workload at each length;
they are not averages over the RULER evaluation set.

Each configuration receives five warm-up iterations followed by
30 measured iterations using CUDA events.
For each measured iteration \(r\), we sum each phase's time over
all 24 GDN layers.
The native core time is
\begin{equation}
T_{\mathrm{native}}^{(r)}
=
T_{\mathrm{Intra,native}}^{(r)}
+
T_{H,\mathrm{native}}^{(r)}
+
T_{O,\mathrm{native}}^{(r)}.
\label{eq:native_latency}
\end{equation}
The \textsc{SparseGDN} core time includes online selection:
\begin{equation}
T_{\mathrm{SparseGDN}}^{(r)}
=
T_{\mathrm{Selector,sparse}}^{(r)}
+
T_{\mathrm{Intra,sparse}}^{(r)}
+
T_{H,\mathrm{sparse}}^{(r)}
+
T_{O,\mathrm{sparse}}^{(r)}.
\label{eq:sparse_latency}
\end{equation}

For \textsc{SparseGDN}, the timed \(H\) phase includes deferred
\(\mathbf{W}\) materialization and the subsequent event scan.
Native \(\mathbf{W}\) construction is included in Intra.
The timed regions exclude GDN input/output projections and surrounding
preprocessing and postprocessing, as well as softmax-attention layers,
MLPs, tokenization, and autoregressive decoding.
The reported results therefore measure GDN-core prefill latency.

\subsection{Main Results}
\label{sec:main_results}

Table~\ref{tab:accuracy_speedup} summarizes RULER scores,
recurrent-event sparsity, and GDN-core speedup for the default
\textsc{SparseGDN} configuration.
Across the evaluation requests, \textsc{SparseGDN} retains approximately
\(29\%\) of recurrent pair--chunk events on average,
corresponding to approximately \(71\%\) event sparsity.

\begin{table*}[t]
  \centering
  \small
  \setlength{\tabcolsep}{3.2pt}
  \caption{
      RULER scores, recurrent-event sparsity, and GDN-core speedup
      for Qwen3.5-4B and Qwen3.5-9B.
      Scores follow the official RULER scoring rules and are reported
      as percentages.
      \(\Delta\) is the arithmetic difference between the Sparse and
      Base columns, computed from unrounded scores.
      Both scores and their differences are expressed on the
      percentage scale.
      Speedup compares native Intra+\(H\)+\(O\) against
      \textsc{SparseGDN} Selector+Intra+\(H\)+\(O\), using the
      24-layer GDN-core timing protocol in
      Section~\ref{sec:experimental_setup}.
      Event sparsity is the mean fraction of valid recurrent
      pair--chunk events omitted across evaluation requests;
      padding chunks are excluded.
      The All row averages across the six context lengths.
  }
  \label{tab:accuracy_speedup}
  \begin{tabular}{lccccc@{\hspace{0.7em}}ccccc}
      \toprule
      & \multicolumn{5}{c}{Qwen3.5-4B}
      & \multicolumn{5}{c}{Qwen3.5-9B} \\
      \cmidrule(lr){2-6}
      \cmidrule(lr){7-11}
      Length
      & Base
      & Sparse
      & \(\Delta\) (\%)
      & Event sp. (\%)
      & Speedup
      & Base
      & Sparse
      & \(\Delta\) (\%)
      & Event sp. (\%)
      & Speedup \\
      \midrule
      4K
      & 96.641 & 96.377 & \(-0.264\) & 70.937 & \(1.498\times\)
      & 96.821 & 96.846 & \(+0.026\) & 70.921 & \(1.500\times\) \\

      8K
      & 96.282 & 96.333 & \(+0.051\) & 70.955 & \(1.504\times\)
      & 96.872 & 96.179 & \(-0.692\) & 71.005 & \(1.506\times\) \\

      16K
      & 96.121 & 96.462 & \(+0.341\) & 70.915 & \(1.696\times\)
      & 96.369 & 96.308 & \(-0.062\) & 70.971 & \(1.702\times\) \\

      32K
      & 96.344 & 96.097 & \(-0.246\) & 70.934 & \(1.830\times\)
      & 96.744 & 96.831 & \(+0.087\) & 70.917 & \(1.825\times\) \\

      64K
      & 95.985 & 96.274 & \(+0.290\) & 70.998 & \(1.891\times\)
      & 96.641 & 96.538 & \(-0.103\) & 70.976 & \(1.892\times\) \\

      128K
      & 95.482 & 95.413 & \(-0.069\) & 70.997
      & \(\mathbf{1.943\times}\)
      & 96.256 & 96.310 & \(+0.054\) & 70.949
      & \(\mathbf{1.929\times}\) \\

      \midrule
      \textbf{All}
      & \textbf{96.142}
      & \textbf{96.159}
      & \(\mathbf{+0.017}\)
      & \textbf{70.956}
      & --
      & \textbf{96.617}
      & \textbf{96.502}
      & \(\mathbf{-0.115}\)
      & \textbf{70.957}
      & -- \\
      \bottomrule
  \end{tabular}
\end{table*}

\paragraph{Long-context accuracy.}
For Qwen3.5-4B, the aggregate RULER score changes from
\(96.142\%\) to \(96.159\%\), a difference of \(+0.017\%\).
The largest absolute score difference across context lengths
is \(0.341\%\) at 16K.

For Qwen3.5-9B, the aggregate score changes from
\(96.617\%\) to \(96.502\%\), a difference of \(-0.115\%\).
The largest score decrease is \(0.692\%\) at 8K.
At each of the other five context lengths, the absolute score
difference is at most \(0.103\%\).
The aggregate scores over 7,800 examples remain close to their
respective baselines for both models.

\paragraph{GDN-core efficiency.}
Both models achieve approximately \(1.5\times\) speedup at 4K,
including online selection overhead.
Across the measured lengths, the total GDN-core speedup increases
with context length, reaching \(1.943\times\) for Qwen3.5-4B
and \(1.929\times\) for Qwen3.5-9B at 128K.

In these measurements, selection overhead accounts for a smaller
fraction of sparse core time at longer contexts.
Longer inputs provide more chunk-level work over which to amortize
fixed kernel-launch overhead, while increasing the absolute savings
from sparse computation and reduced state traffic.
The similar speedups across the two models are consistent with their
identical GDN-core tensor geometry and number of GDN layers.

\begin{figure}[t]
    \centering
    \includegraphics[width=\linewidth]{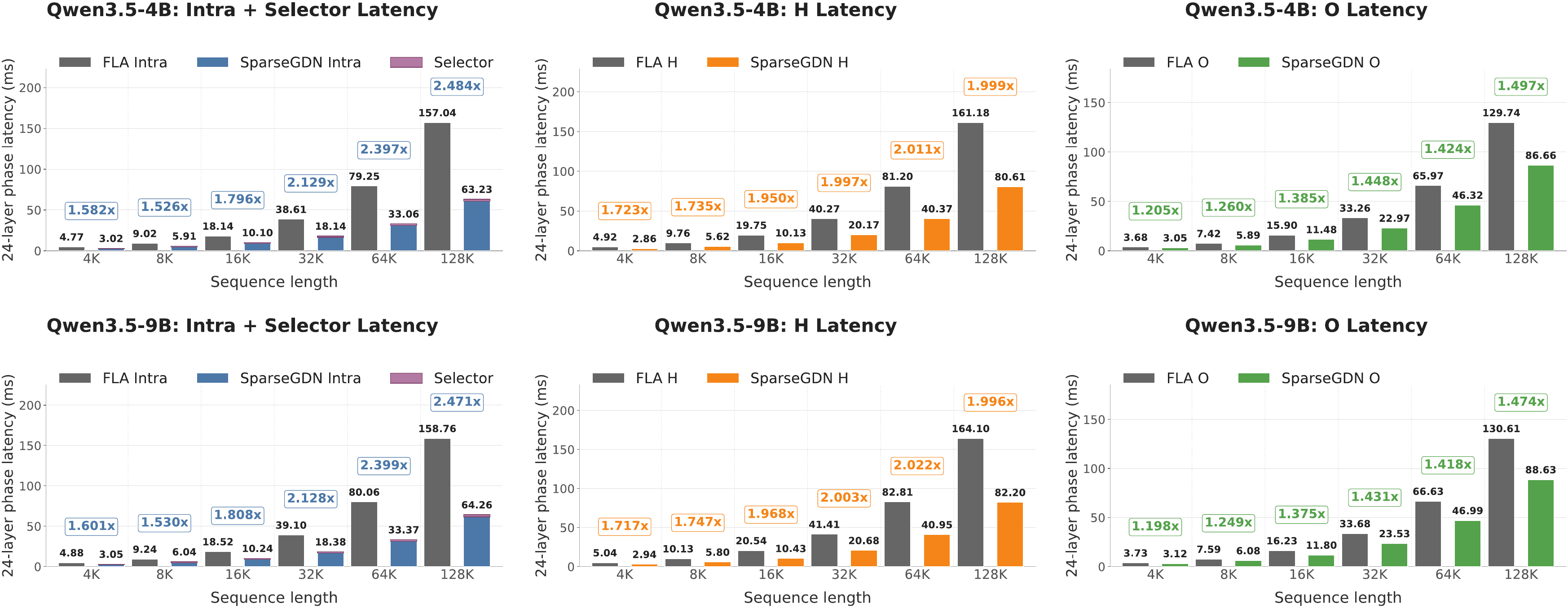}
    \caption{
      \textbf{Per-stage GDN-core prefill latency.}
      The top and bottom rows show Qwen3.5-4B and Qwen3.5-9B,
      respectively.
      The left column compares native Intra with
      \textsc{SparseGDN} Intra plus online Selector;
      the middle and right columns compare \(H\) and \(O\).
      For \textsc{SparseGDN}, \(H\) includes deferred
      \(\mathbf{W}\) materialization.
      Each component reports the median of per-iteration CUDA-event
      times summed over all 24 GDN layers. 
      The stacked Intra-plus-Selector bars sum the two component medians.
      Boxed values denote the corresponding native-to-sparse latency
      ratios. Lower latency is better.
    }
    \label{fig:phase_latency_breakdown}
\end{figure}

\paragraph{Stage-wise latency breakdown.}
Figure~\ref{fig:phase_latency_breakdown} separates the core computation
into Intra plus online Selector, recurrent-state propagation (\(H\)),
and output reconstruction (\(O\)).
Across the evaluated context lengths, the combined Intra-plus-Selector
phase achieves \(1.53\times\)--\(2.48\times\) speedup over native Intra
for Qwen3.5-4B and \(1.53\times\)--\(2.47\times\) for Qwen3.5-9B.
This reduction reflects the block-adjacent solve, progressive value
computation, and the relocation of \(\mathbf{W}\) construction
to the timed \(H\) phase, where it is performed only for retained events.

Including deferred \(\mathbf{W}\) materialization, the \(H\) phase
achieves approximately \(1.72\times\) speedup at 4K and
\(2.00\times\) at 128K for both models.
The event scan omits recurrent read corrections and state writes
for skipped chunks while preserving their accumulated decay.

The \(O\) phase achieves \(1.20\times\)--\(1.50\times\) speedup
across the two models and evaluated lengths.
Its gains are more modest because outputs and local causal interactions
are still computed for every token.
Savings arise from compact state snapshots and mapped group-wise
reconstruction, which reuse a group-entry state across two adjacent
chunks.

\subsection{True-Sparse Batch Scaling}
\label{sec:batch_scaling}

We extend the batch-size-one evaluation to batch sizes 2, 4, and 8.
The selector, event masks, state layouts, and \(H/O\) kernels
operate with an explicit batch dimension.
For batch size \(B\), we use the first \(B\) requests from
\texttt{niah\_single\_1}, repeating or truncating each tokenized
prompt to the same target length.
Native and \textsc{SparseGDN} receive identical batch inputs.

\begin{figure*}[t]
  \centering
  \includegraphics[width=\textwidth]{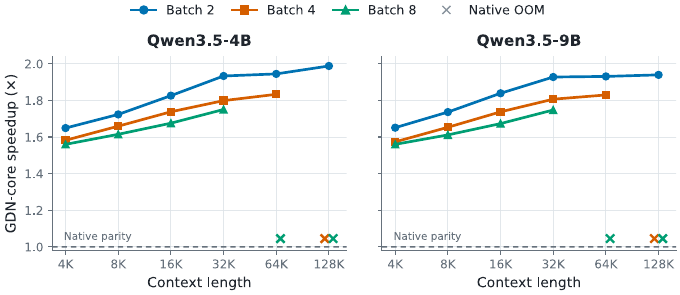}
  \caption{
      \textbf{GDN-core speedup across batch sizes and context lengths.}
      Speedup is Native/Sparse, includes online selection, and uses
      batch latencies aggregated over all 24 GDN layers.
      Crosses mark configurations without a paired speedup because
      Native runs out of memory.
      Crosses are placed at a fixed height for display and do not
      represent measured speedups.
      The corresponding sparse-only outcomes are detailed in
      Table~\ref{tab:true_sparse_batch}.
      Every completed paired configuration favors \textsc{SparseGDN}.
      Higher is better.
  }
  \label{fig:true_sparse_batch_scaling}
\end{figure*}

For this experiment, CUDA events measure a contiguous core interval
in each GDN layer.
The native interval covers Intra, \(H\), and \(O\), while the sparse
interval additionally includes online selection.
Deferred \(\mathbf{W}\) materialization is included in the sparse
\(H\) phase.
We use one warm-up iteration followed by five measured iterations.
For each iteration, we sum the intervals over all 24 GDN layers
and report the median of these totals.
These are GDN-core latencies for an entire batch.
Speedup is the native median divided by the sparse median.
Figure~\ref{fig:true_sparse_batch_scaling} shows the scaling trends,
and Table~\ref{tab:true_sparse_batch} reports the corresponding
speedups and memory-related missing results.

\begin{table*}[t]
  \centering
  \small
  \setlength{\tabcolsep}{5pt}
  \caption{
      Batched GDN-core speedup (Native/Sparse).
      \(\mathrm{OOM}^{\dagger}\) indicates that Native runs out of
      memory in the benchmark harness, while \textsc{SparseGDN}
      completes in a separate process.
      Plain \(\mathrm{OOM}\) indicates that Native runs out of memory
      and sparse completion has not been independently verified.
      Speedups include online selection and cover only the GDN core.
  }
  \label{tab:true_sparse_batch}
  \begin{tabular}{lccccccc}
      \toprule
      Model & Batch & 4K & 8K & 16K & 32K & 64K & 128K \\
      \midrule
      4B & 2
      & \(1.649\times\) & \(1.723\times\) & \(1.826\times\)
      & \(1.933\times\) & \(1.945\times\) & \(1.988\times\) \\
      4B & 4
      & \(1.582\times\) & \(1.659\times\) & \(1.738\times\)
      & \(1.799\times\) & \(1.833\times\) & \(\mathrm{OOM}^{\dagger}\) \\
      4B & 8
      & \(1.560\times\) & \(1.615\times\) & \(1.675\times\)
      & \(1.750\times\) & \(\mathrm{OOM}^{\dagger}\) & \(\mathrm{OOM}\) \\
      \midrule
      9B & 2
      & \(1.651\times\) & \(1.736\times\) & \(1.839\times\)
      & \(1.928\times\) & \(1.931\times\) & \(1.939\times\) \\
      9B & 4
      & \(1.574\times\) & \(1.653\times\) & \(1.737\times\)
      & \(1.807\times\) & \(1.830\times\) & \(\mathrm{OOM}\) \\
      9B & 8
      & \(1.560\times\) & \(1.611\times\) & \(1.673\times\)
      & \(1.748\times\) & \(\mathrm{OOM}\) & \(\mathrm{OOM}\) \\
      \bottomrule
  \end{tabular}
\end{table*}

Across all completed paired configurations, \textsc{SparseGDN}
achieves \(1.560\times\)--\(1.988\times\) GDN-core speedup.
At a fixed batch size, speedup increases with context length
across the measured points.
At a fixed length, the relative speedup decreases as batch size grows.
For example, at 32K, increasing batch size from 2 to 8 changes
speedup from \(1.933\times\) to \(1.750\times\) for Qwen3.5-4B
and from \(1.928\times\) to \(1.748\times\) for Qwen3.5-9B.
Across the completed comparisons, the native baseline's per-sample
latency decreases more rapidly with increasing batch size than
that of \textsc{SparseGDN}, narrowing the relative speedup.

At the missing long-context points, the native execution runs out
of memory in the benchmark harness before a paired timing result
can be obtained.
Separate sparse-only runs complete the Qwen3.5-4B batch-4/128K
and batch-8/64K configurations in 843.411 ms and 835.367 ms,
respectively.
These measurements establish sparse-path feasibility for the two
configurations; speedups remain unavailable because the corresponding
native runs did not complete.
For the remaining OOM entries, sparse completion has not been
independently verified.

\subsection{Accuracy--Efficiency Trade-off}
\label{sec:retention_tradeoff}

The default operating point retains approximately \(29\%\) of recurrent
events. We explore the accuracy--efficiency trade-off by varying the
low-, medium-, and high-tier retention rates while keeping the selector
rules and kernel pipeline unchanged. The five settings of
\((\rho_{\mathrm L},\rho_{\mathrm M},\rho_{\mathrm H})\) are
\((0.028,0.10,0.317)\), \((0.056,0.20,0.633)\),
\((0.08,0.30,0.95)\), \((0.20,0.40,1.0)\), and
\((0.333,0.50,1.0)\), corresponding to the 10\%-budget,
20\%-budget, default, 40\%-budget, and 50\%-budget configurations,
respectively. These budget labels denote nominal targets; the reported
keep rates are measured over evaluation requests. Head-pair assignments
and selected chunks are recomputed for each input at each operating point.

\begin{figure*}[t]
  \centering
  \includegraphics[width=\textwidth]{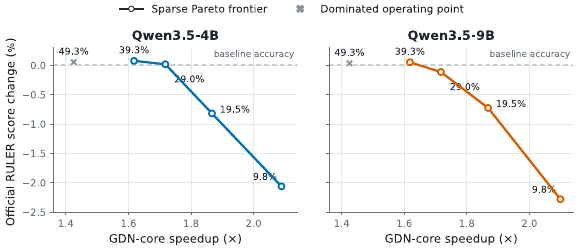}
  \caption{
      \textbf{Accuracy--efficiency trade-off across retention budgets.}
      Each panel plots the aggregate RULER score difference
      \(\Delta\), defined in Table~\ref{tab:retention_pareto}, against
      the geometric mean GDN-core speedup over 4K--128K.
      Point labels show measured recurrent-event keep rates.
      Solid lines connect nondominated configurations among the
      evaluated sparse operating points. The cross marks the
      approximately \(49\%\)-keep configuration, which is dominated
      by the approximately \(39\%\)-keep configuration for both models
      under these aggregate metrics. Speedups include online selection.
      Higher and farther right is better.
  }
  \label{fig:retention_pareto}
\end{figure*}

Table~\ref{tab:retention_pareto} summarizes the aggregate results, and
Figure~\ref{fig:retention_pareto} plots the evaluated operating points.
All speedups in this section are geometric means over the six context
lengths from 4K to 128K, measured at batch size one with online selection
included. Reducing mean keep to approximately \(9.8\%\) yields speedups
of about \(2.1\times\), with aggregate score changes of \(-2.063\%\)
and \(-2.281\%\) for Qwen3.5-4B and Qwen3.5-9B, respectively.
The default configuration provides \(1.718\times\) and \(1.717\times\)
speedups with score changes of \(+0.017\%\) and \(-0.115\%\).
At approximately \(39.3\%\) keep, both models achieve a
\(1.618\times\) speedup and slightly higher observed aggregate scores
than their corresponding baselines, with changes of \(+0.075\%\)
and \(+0.051\%\).

The approximately \(49\%\)-keep configuration is slower and has a
slightly lower aggregate score than the approximately \(39\%\)-keep
configuration for both models. It is therefore dominated among the
evaluated operating points under these aggregate metrics. This comparison
does not imply the same ordering at every context length.

\begin{table*}[t]
  \centering
  \small
  \setlength{\tabcolsep}{5pt}
  \caption{
      Retention-rate trade-off over all six context lengths.
      Keep is the mean fraction of valid recurrent pair--chunk events
      retained across evaluation requests; padding chunks are excluded.
      \(\Delta = S_{\mathrm{sparse}}-S_{\mathrm{base}}\) is the arithmetic
      difference in aggregate RULER scores, computed from unrounded
      scores over 7,800 examples per model using the official scoring
      rules. Scores and their differences are expressed on the
      percentage scale. Speedup is the geometric mean of the six
      GDN-core speedups at batch size one under the timing protocol
      in Section~\ref{sec:experimental_setup}, including online selection.
      Higher speedup and \(\Delta\) are better.
  }
  \label{tab:retention_pareto}
  \begin{tabular}{lccc@{\hspace{1.2em}}ccc}
      \toprule
      & \multicolumn{3}{c}{Qwen3.5-4B}
      & \multicolumn{3}{c}{Qwen3.5-9B} \\
      \cmidrule(lr){2-4}
      \cmidrule(lr){5-7}
      Operating point
      & Keep (\%) & \(\Delta\) (\%) & Speedup
      & Keep (\%) & \(\Delta\) (\%) & Speedup \\
      \midrule
      10\%-budget
      & 9.760 & \(-2.063\) & \(2.089\times\)
      & 9.755 & \(-2.281\) & \(2.099\times\) \\
      20\%-budget
      & 19.493 & \(-0.821\) & \(1.867\times\)
      & 19.502 & \(-0.727\) & \(1.868\times\) \\
      Default
      & 29.044 & \(+0.017\) & \(1.718\times\)
      & 29.043 & \(-0.115\) & \(1.717\times\) \\
      40\%-budget
      & 39.253 & \(+0.075\) & \(1.618\times\)
      & 39.261 & \(+0.051\) & \(1.618\times\) \\
      50\%-budget
      & 49.325 & \(+0.053\) & \(1.425\times\)
      & 49.337 & \(+0.033\) & \(1.425\times\) \\
      \bottomrule
  \end{tabular}
\end{table*}

\subsection{Algorithmic and Selector Ablations}
\label{sec:algorithmic_ablations}

Table~\ref{tab:algorithmic_ablation} compares intra-chunk approximations and
recurrent-event policies on the complete 4K RULER set. All custom variants
use the same group-local output rule from Section~\ref{sec:mapped_o}.

The adaptive variants apply sparse H corrections and writes while retaining
group-local reads: dropped even chunks still contribute to the
following odd chunk's output as specified by
Equation~\eqref{eq:group2_local_read_difference}. These comparisons do not
separately measure the effect of that local contribution. Full64 uses a
custom BF16 kernel and is not bitwise identical to native FLA. Moreover,
each adaptive variant recomputes its mask from its own Intra outputs;
comparing Full64+adaptive with \textsc{SparseGDN} therefore includes mask
reselection as well as the change in Intra.

\begin{table*}[t]
  \centering
  \small
  \setlength{\tabcolsep}{4.5pt}
  \caption{
      Algorithmic component ablation on all 13 RULER tasks at 4K
      (1,300 examples per model). Scores and their arithmetic differences
      are expressed on the percentage scale. \(\Delta\) is the score
      difference from Native, computed from unrounded scores.
      Full64 + adaptive restores the full lower-triangular 64-token
      intra-chunk solve while retaining adaptive event selection;
      Block-adjacent + all retains every recurrent event.
      Adaptive and random variants use matched approximately
      \(29\%\) event budgets.
      All custom variants retain the same group-local output rule;
      Native uses the dense operator. Adaptive masks are recomputed for
      each Intra variant. Bold highlights the proposed method.
  }
  \label{tab:algorithmic_ablation}
  \begin{tabular}{lccc@{\hspace{1.0em}}ccc}
      \toprule
      & \multicolumn{3}{c}{Qwen3.5-4B}
      & \multicolumn{3}{c}{Qwen3.5-9B} \\
      \cmidrule(lr){2-4}
      \cmidrule(lr){5-7}
      Variant
      & Score & \(\Delta\) (\%) & Keep (\%)
      & Score & \(\Delta\) (\%) & Keep (\%) \\
      \midrule
      Native
      & 96.641 & \(+0.000\) & 100.000
      & 96.821 & \(+0.000\) & 100.000 \\
      Block-adjacent + all
      & 96.410 & \(-0.231\) & 100.000
      & 96.846 & \(+0.026\) & 100.000 \\
      Full64 + adaptive
      & 96.669 & \(+0.028\) & 28.991
      & 96.787 & \(-0.033\) & 29.017 \\
      Block-adjacent + random
      & 95.067 & \(-1.574\) & 29.029
      & 95.621 & \(-1.200\) & 29.029 \\
      \textbf{\textsc{SparseGDN}}
      & 96.377 & \(-0.264\) & 29.063
      & 96.846 & \(+0.026\) & 29.079 \\
      \bottomrule
  \end{tabular}
\end{table*}

The matched-budget random baseline confirms that accuracy does not follow
from sparsity alone. At 4K, adaptive selection exceeds random selection by
\(1.310\%\) for 4B (paired bootstrap 95\% CI
\([0.481,2.154]\%\)) and \(1.226\%\) for 9B
(\([0.494,1.983]\%\)). Across all six lengths, the aggregate gaps grow to
\(2.006\%\) and \(1.614\%\), respectively; adaptive selection wins in
all 12 model--length comparisons at essentially identical mean keep.

We further probe the selector components using a task-stratified set of
130 8K requests per model. Dense-decode importance estimates each prompt
chunk's contribution along the native dense decoding trajectory.
Non-anchor importance-mass recall measures the fraction of this estimated
importance retained among non-anchor events. Recall and its arithmetic
differences are expressed on the percentage scale. We disable output-gate
visibility in the selector for these diagnostics to isolate the roles of
tail-based ranking and recurrent lifetime; the model's output gate remains
unchanged. Adding recurrent lifetime to the tail statistic raises
non-anchor importance-mass recall by \(32.037\%\) for 4B and
\(31.484\%\) for 9B. Holding head-pair tiers and exact budgets fixed,
tail-based chunk ranking outperforms random ranking by \(14.171\%\)
and \(13.646\%\). Both effects hold on all 130 requests for each model.

In a separate end-to-end 8K ablation over 1,300 examples per model, we set
the selector's output-gate visibility factor to one while retaining the
model's output gate. This yields observed score decreases of \(0.487\%\)
and \(0.410\%\) for 4B and 9B, respectively, despite a slight increase
in mean keep. The paired 95\% bootstrap confidence intervals for the score
change after removing visibility are \([-1.128,+0.128]\%\) and
\([-1.077,+0.205]\%\), respectively; both include zero.
The proxy-recall results diagnose routing and ranking; they are not
substitutes for the end-to-end RULER comparisons above.

\paragraph{Adaptive tier counts.}
We evaluate the benefit of adapting tier counts in
Section~\ref{sec:selector} by fixing
\((n_{\mathrm L},n_{\mathrm M},n_{\mathrm H})=(9,4,3)\) in every GDN
layer. The fixed variant selects three high-tier pairs first, then four
medium-tier pairs from the remainder, leaving nine low-tier pairs.
Pair identities remain input-dependent. Score computation, pair and chunk
ranking, the \((0.08,0.30,0.95)\) retention rates, tie-breaking, valid-chunk
quotas, anchors, and the sparse execution path are unchanged. We evaluate
this variant on the same 7,800 Qwen3.5-9B examples as the main evaluation;
the dynamic results reuse the corresponding outputs in
Table~\ref{tab:accuracy_speedup}.

Table~\ref{tab:adaptive_tier_counts} shows that dynamic allocation raises
the aggregate official score from \(96.183\%\) to \(96.502\%\), while
reducing mean event keep from \(29.816\%\) to \(29.043\%\). The paired gain
is \(0.319\%\) (95\% bootstrap CI \([0.114,0.532]\%\)).
We use 20,000 paired resamples stratified by task and context length;
intervals are pointwise, without multiple-comparison adjustment.
The largest gain occurs at 128K (\(+1.044\%\)), although fixed counts
perform slightly better at 8K (\(+0.256\%\)). These results support
adapting tier counts to improve the accuracy--retention trade-off on
Qwen3.5-9B. Actual retention is reported separately for the two variants;
this ablation does not measure selector latency.

\begin{table*}[t]
  \centering
  \small
  \setlength{\tabcolsep}{3.2pt}
  \caption{
      Effect of adaptive tier counts on Qwen3.5-9B. Scores are official
      RULER percentages over 100 examples per task at each context length
      (13 tasks; 7,800 examples in total). All is the mean over the six
      lengths. Score differences reported in the text are computed from
      unrounded values.
      Keep is the mean per-request fraction of valid recurrent pair--chunk
      events retained, expressed as a percentage. Both variants use the
      same retention tiers; realized keep rates are not forced to match.
      Bold indicates the higher score at each length and in aggregate.
  }
  \label{tab:adaptive_tier_counts}
  \begin{tabular}{lcccccccc}
      \toprule
      Allocation & 4K & 8K & 16K & 32K & 64K & 128K & All & Keep (\%) \\
      \midrule
      Fixed counts \(9/4/3\)
      & 96.691 & \textbf{96.436} & 95.815 & 96.505 & 96.385 & 95.267
      & 96.183 & 29.816 \\
      Dynamic counts
      & \textbf{96.846} & 96.179 & \textbf{96.308} & \textbf{96.831}
      & \textbf{96.538} & \textbf{96.310} & \textbf{96.502} & 29.043 \\
      \bottomrule
  \end{tabular}
\end{table*}

\paragraph{Tail-window selection.}
We use Tail-16 as the default ranking window in
Equation~\ref{eq:tail_score}, balancing ranking quality with score-computation
cost. To isolate the window's effect on ranking, we fix the Tail-16
pair-tier assignments, exact per-pair keep counts, and first/last-chunk
anchors, and vary only the score used to order non-anchor chunks. The
recall diagnostic uses 13 RULER tasks with 10 stratified 8K requests per
task and model, under production routing with output-gate visibility.
For each request, we average over all 24 GDN layers before performing
20,000 task-stratified bootstrap resamples.

\begin{figure*}[t]
  \centering
  \includegraphics[width=\textwidth]{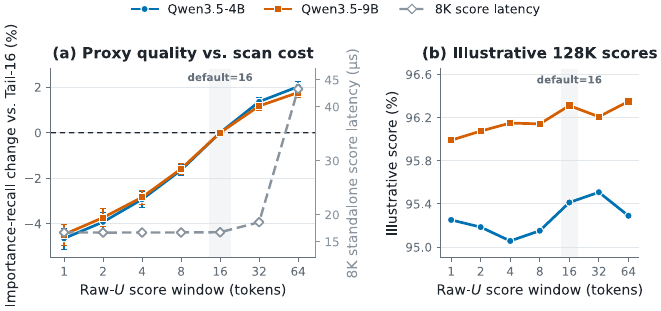}
  \caption{
      \textbf{Tail-window ranking quality, scan cost, and illustrative
      task scores.}
      \textbf{Left:} change in retained non-anchor dense-decode importance
      mass relative to Tail-16 on 130 stratified 8K requests per model,
      with task-stratified paired 95\% bootstrap confidence intervals.
      Pair-tier assignments, exact per-pair keep counts, and anchors
      are fixed to Tail-16. The gray curve reports standalone
      score-kernel latency per layer at 8K.
      \textbf{Right:} illustrative 128K score curves.
      The shaded bands highlight the default Tail-16 window in both panels.
  }
  \label{fig:tail_window_ablation}
\end{figure*}

The left panel of Figure~\ref{fig:tail_window_ablation} shows the recall
lost when the ranking window is shortened. Moving from Tail-8 to
Tail-16 raises importance-mass recall by \(1.643\%\) and \(1.590\%\)
for 4B and 9B, respectively. Under the same production routing condition,
Tail-16 exceeds random ordering by \(14.345\%\) and \(14.098\%\).
Longer windows retain additional diagnostic importance:
Tail-32 improves over Tail-16 by \(1.362\%\) and \(1.172\%\), while
Tail-64 improves it by \(2.024\%\) and \(1.764\%\). Recall alone thus
favors increasing the window, so the default must also account for the
cost of computing that score.

The gray curve makes this cost explicit. On 8K, the standalone
score kernel takes \(16.705\), \(16.739\), \(18.606\), and \(43.291\)
microseconds per layer for Tail-8, Tail-16, Tail-32, and Tail-64,
respectively. Tail-16 recovers more importance than Tail-8 at nearly the
same scan cost; Tail-64 costs approximately \(2.59\times\) as much as
Tail-16. These timings measure a standalone reread of the materialized
\(\mathbf{U}\) tensor. In the production pipeline, the Tail-16 reduction
is fused into Intra while the final 16-token tile \(\mathbf{UB}_3\) is
resident on chip. Both Tail-8 and Tail-16 fit in this tile, whereas
Tail-32 and Tail-64 require accumulation across earlier tiles. Thus the
independent scan measurements characterize score cost, rather than a
measured GDN-core speedup between window choices.

In this plotted comparison, Tail-16 reaches \(95.413\%\) and \(96.310\%\)
for 4B and 9B, respectively, matching Table~\ref{tab:accuracy_speedup}.
Relative to the illustrative Tail-8 values of \(95.151\%\) and
\(96.141\%\), the plotted increases are \(0.262\%\) and \(0.169\%\),
respectively. Extending the window beyond 16 tokens does not improve
both models consistently: Tail-32 increases the 4B score to \(95.508\%\),
but lowers the 9B score to \(96.208\%\); Tail-64 reaches \(95.290\%\)
and \(96.349\%\), respectively, at substantially higher scan cost.
These illustrative comparisons do not establish statistical significance.

We therefore retain Tail-16 as the default. It captures more diagnostic
importance than Tail-8 at nearly the same standalone score cost and fits
the final tile already used by the fused Intra kernel. Longer windows
provide additional recall but require more score accumulation, and the
task-score curves offer no consistent advantage that grows with window
length. Tail-16 is a practical balance between ranking quality and
execution cost, rather than the window with the highest recall or the
highest score for every model.

\subsection{Kernel-Pipeline Ablations}
\label{sec:kernel_ablations}

Table~\ref{tab:kernel_ablation} evaluates the execution mechanisms that
reduce GPU stage latency under the same retained masks and group-local
output rule. Each optimization is compared with a reference matched to
its producer and consumer formats. On the benchmark inputs, all variants
match the production retained masks across all 24 GDN layers and produce
bitwise-identical prefill logits at the last input token.

The Deferred-\(\mathbf{W}\) comparison replaces full \(\mathbf{W}\)
materialization with deferred materialization for retained events,
including the associated packed-\(\mathbf{W}\) consumer path.
The CSR reference performs recurrent GEMMs at every chunk while preserving
the same event mask, prefix-gap decay arithmetic, and compact output
layout; the optimized kernel scans only retained events.
The compact-state comparison measures the cost of explicitly expanding
compact event states into dense group-entry states after CSR propagation,
with mapped \(O\) fixed in both variants.

For the isolated mapped-\(O\) comparison, both variants materialize the
same dense group-entry states in \(H\), and only \(O\) is timed.
The reference reads these materialized states, whereas mapped \(O\)
reconstructs them from compact event snapshots. Both output paths evaluate
the two-chunk local reads in Section~\ref{sec:mapped_o}, including the
contribution of a dropped even chunk. The joint Compact+mapped comparison
also removes dense-state materialization and measures the combined effect
over Intra+\(H\)+\(O\).

Compact states yield geometric-mean \(H\)-stage speedups of approximately
\(1.40\times\), while deferred \(\mathbf{W}\) yields approximately
\(1.34\times\) over Intra+\(H\)+\(O\).
At 128K, CSR event scanning and mapped output reconstruction reach
approximately \(1.23\times\) and \(1.37\times\), respectively.
Tier-aware scheduling is enabled at 16K and above, where it provides
\(1.04\)--\(1.05\times\) speedups. At 4K and 8K, both scheduling variants
use the same execution path; these lengths are also included in the
geometric means reported in the table.

\begin{table*}[t]
  \centering
  \small
  \setlength{\tabcolsep}{4.5pt}
  \caption{
      Kernel-pipeline ablations at batch size one on a GPU.
      Speedup is the ratio of reference to optimized median latency
      over the stages listed in Scope. Each variant uses five warmup
      forwards and 30 interleaved CUDA-event measurements. Within each
      measurement, the indicated stage times are summed across all
      24 GDN layers before computing the median across measurements.
      Separately timed Selector work is excluded; statistics fused into
      Intra remain included. Geomean is the geometric mean of speedups
      over the six lengths from 4K to 128K; the 128K column reports the
      longest-context result. Each row uses its own reference, so the
      isolated speedups should not be multiplied. Compact+mapped reports
      the joint effect of compact state storage and mapped output reads.
  }
  \label{tab:kernel_ablation}
  \begin{tabular}{llcc@{\hspace{1.0em}}cc}
      \toprule
      & & \multicolumn{2}{c}{Qwen3.5-4B}
      & \multicolumn{2}{c}{Qwen3.5-9B} \\
      \cmidrule(lr){3-4}
      \cmidrule(lr){5-6}
      Optimization & Scope
      & Geomean & 128K
      & Geomean & 128K \\
      \midrule
      Deferred-\(\mathbf{W}\) & Intra+\(H\)+\(O\)
      & \(1.336\times\) & \(1.390\times\)
      & \(1.338\times\) & \(1.391\times\) \\
      CSR event scan & \(H\)
      & \(1.149\times\) & \(1.226\times\)
      & \(1.153\times\) & \(1.230\times\) \\
      Compact states & \(H\)
      & \(1.398\times\) & \(1.406\times\)
      & \(1.402\times\) & \(1.405\times\) \\
      Tier scheduling & \(H\)
      & \(1.029\times\) & \(1.044\times\)
      & \(1.030\times\) & \(1.045\times\) \\
      Mapped \(O\) & \(O\)
      & \(1.239\times\) & \(1.374\times\)
      & \(1.226\times\) & \(1.368\times\) \\
      Compact + mapped & Intra+\(H\)+\(O\)
      & \(1.226\times\) & \(1.285\times\)
      & \(1.224\times\) & \(1.283\times\) \\
      \bottomrule
  \end{tabular}
\end{table*}

The similar ablation trends across the two models are consistent with
their shared 24-layer GDN geometry. These results show that reducing update
materialization and state traffic contributes to the measured speedups
alongside sparse recurrent execution. 

\subsection{Peak-Memory Scaling}
\label{sec:peak_memory}

We measure peak allocated memory during complete-model BF16 prefill
with batch size one, \texttt{use\_cache=True}, and logits retained only
for the last input token (\texttt{logits\_to\_keep=1}).
We use the same repeated-or-truncated RULER prompt from 4K through 256K.
Each configuration uses one warmup forward followed by three measured
forwards. Before each measurement, we release unused cached allocator
blocks and reset the peak-memory statistics. For completed configurations,
we report the maximum across the three measurements.

\begin{figure*}[t]
  \centering
  \includegraphics[width=\textwidth]{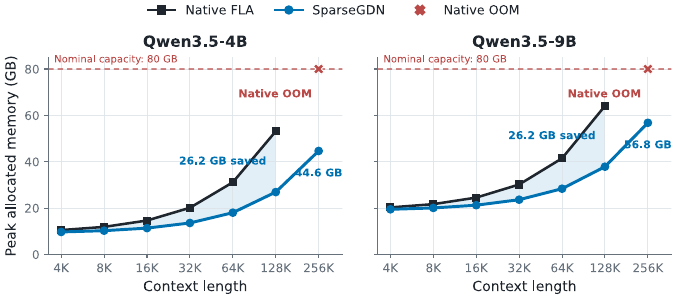}
  \caption{
      \textbf{Complete-model prefill peak allocated memory across
      context lengths.}
      Measurements use BF16, batch size one, \texttt{use\_cache=True}.
      The shaded region shows the absolute memory saved by
      \textsc{SparseGDN} wherever both paths complete.
      The dashed line marks the nominal 80 GB device capacity as a
      reference. Crosses indicate Native OOM at 256K and are placed
      on this reference line for display; they do not represent
      measured allocation peaks. \textsc{SparseGDN} completes at
      that length for both models. Lower is better.
  }
  \label{fig:peak_memory_scaling}
\end{figure*}

Figure~\ref{fig:peak_memory_scaling} shows the peak-memory curves, while
Table~\ref{tab:peak_memory} reports the 128K and 256K results. The forward increment is
\(M_{\mathrm{incr}}=M_{\mathrm{peak}}-M_{\mathrm{entry}}\), where
\(M_{\mathrm{entry}}\) is the allocation already live immediately before
the forward. This subtraction excludes persistent allocations such as
model weights from the incremental measurement.

At 128K, \textsc{SparseGDN} reduces the absolute peak from
53.116 to 26.894 GB for Qwen3.5-4B and from 64.066 to 37.844 GB
for Qwen3.5-9B. Both models save 26.222 GB, consistent with their
shared GDN tensor geometry. The corresponding forward-increment
reductions are \(59.56\%\) and \(57.97\%\), respectively.

\begin{table*}[t]
  \centering
  \small
  \setlength{\tabcolsep}{5pt}
  \caption{
      Complete-model prefill peak allocated memory. ``Incr.'' denotes the peak allocation above memory already
      live at forward entry. ``Incr. reduction'' is the relative
      decrease in this increment, computed from unrounded values.
      Native OOMs during its first 256K warmup, so no completed-forward
      peak or increment is available. Lower is better.
  }
  \label{tab:peak_memory}
  \begin{tabular}{lcccccc}
      \toprule
      Model & Length & Native peak & Sparse peak
      & Native incr. & Sparse incr. & Incr. reduction \\
      \midrule
      4B & 128K & 53.116 & 26.894 & 44.027 & 17.804 & 59.56\% \\
      9B & 128K & 64.066 & 37.844 & 45.235 & 19.012 & 57.97\% \\
      4B & 256K & OOM & 44.637 & OOM & 35.544 & -- \\
      9B & 256K & OOM & 56.794 & OOM & 37.960 & -- \\
      \bottomrule
  \end{tabular}
\end{table*}

Across all completed Native comparisons from 4K to 128K, the forward
increment decreases by \(58.36\%\)--\(59.56\%\) for 4B and
\(56.83\%\)--\(57.97\%\) for 9B.
At the models' configured maximum context length of 256K,
both native FLA runs encounter OOM during warmup, whereas
\textsc{SparseGDN} completes with absolute allocated peaks of
44.637 and 56.794 GB for 4B and 9B, respectively.
These 256K results establish prefill feasibility under the tested
configuration; they do not provide quantitative Native peak-memory
comparisons.

\section{Conclusion and Limitations}
\label{sec:conclusion}

We presented \textsc{SparseGDN}, a hardware-aware framework that accelerates long-context GDN prefill by combining a block-adjacent intra-chunk approximation with input-dependent recurrent-event selection and a co-designed sparse GPU pipeline. Deferred update materialization, CSR-based
state propagation, compact event states, and mapped output reconstruction translate approximately \(71\%\) recurrent-event sparsity into 1.50-1.94$\times$ GDN-core speedups on Qwen3.5-4B and Qwen3.5-9B, while preserving comparable aggregate RULER accuracy from 4K to 128K tokens. 

The main limitation lies in the selector: it relies on hand-designed tail, lifetime, and output-gate statistics, chooses among fixed retention tiers, and shares each decision across an
adjacent value-head pair. Although this design is efficient and performs well in our experiments, it is not learned end to end and may miss rare but important state transitions or require recalibration for different models, tasks, and data distributions. Evaluation is also limited to RULER, whose
largely controlled retrieval and reasoning tasks do not fully represent open-ended generation or real application workloads; moreover, aggregate scores can conceal regressions on individual tasks or context lengths. Broader validation on complementary long-context benchmarks and deployment
traces is therefore needed.
\section{Related Work}
\label{sec:related_work}

\subsection{Efficient Long-Context Modeling}

The quadratic cost of self-attention has motivated recurrence, structured
sparsity, low-rank projection, hashing, routing, and kernelized
approximations~\citep{vaswani2017attention,dai2019transformerxl,
child2019sparse,kitaev2020reformer,beltagy2020longformer,
zaheer2020bigbird,wang2020linformer,choromanski2021performer,
roy2021routing}. Long Range Arena benchmarks several families of efficient
sequence models~\citep{tay2021longrange}. Complementary systems work
accelerates exact attention through IO-aware
kernels~\citep{dao2022flashattention,dao2024flashattention2,
shah2024flashattention3} and distributes attention computation across
devices~\citep{liu2023ringattention,brandon2023stripedattention}.
MInference identifies attention-head sparsity patterns offline and
constructs sparse indices dynamically during
prefill~\citep{jiang2024minference}. Native Sparse Attention uses a
trainable hierarchical sparsity pattern, while SpargeAttention applies
online filtering at inference time~\citep{yuan2025nsa,
zhang2025spargeattention}. \textsc{SparseGDN} targets recurrent
head-pair--chunk events within pretrained GDNs while producing outputs
for every token.

\subsection{Linear Attention and Recurrent Models}

Linear attention admits a recurrent form and can be interpreted as a
fast-weight memory~\citep{katharopoulos2020transformers,schlag2021linear}.
Gated Linear Attention, parallel Delta-rule algorithms, and Gated Delta
Networks improve the selectivity and hardware efficiency of this
family~\citep{yang2024gla,yang2024deltarule,yang2024gateddeltanet}.
DeltaProduct improves state tracking through products of Householder
updates, while FG\({}^{2}\)-GDN extends the scalar Delta-rule learning
rate to channel-wise control~\citep{siems2025deltaproduct,sun2026fg2gdn}.

S4, S5, S4D, and the Linear Recurrent Unit explore structured or diagonal
recurrent dynamics~\citep{gu2022efficiently,smith2023s5,gu2022diagonal,
orvieto2023lru}. H3, Hyena, RetNet, RWKV, Mamba, and Mamba-2 further
explore state-space, convolutional, and recurrent sequence
models~\citep{fu2023h3,poli2023hyena,sun2023retnet,peng2023rwkv,
gu2024mamba,dao2024transformers}. Griffin, Jamba, and Samba combine
recurrent or state-space layers with attention, with Jamba also
incorporating mixture-of-experts layers~\citep{de2024griffin,
lieber2025jamba,ren2025samba}. xLSTM revisits gated recurrent memory,
while MEGA combines moving-average representations with gated
attention~\citep{beck2024xlstm,ma2023mega}.
\textsc{SparseGDN} focuses on approximating and sparsely executing
an existing GDN prefill operator without retraining.

\subsection{Dynamic Sparsity and Cache Selection}

Skip RNN learns to omit redundant recurrent
updates~\citep{campos2018skiprnn}. Mixture-of-experts models route tokens
to subsets of parameters and rely on efficient sparse
execution~\citep{shazeer2017moe,lepikhin2021gshard,fedus2022switch,
gale2023megablocks}. Deja Vu and PowerInfer exploit input-dependent
activation sparsity~\citep{liu2023dejavu,song2024powerinfer}, while
SparseGPT and Wanda prune model
weights~\citep{frantar2023sparsegpt,sun2024wanda}.
\textsc{SparseGDN} applies input-dependent sparsity to recurrent events
at the head-pair--chunk level while keeping pretrained parameters fixed.

H2O, Scissorhands, and SnapKV select cached tokens using attention-derived
importance~\citep{zhang2023h2o,liu2023scissorhands,li2024snapkv}.
StreamingLLM retains attention sinks and recent
tokens~\citep{xiao2024streamingllm}; PyramidKV allocates layer-wise cache
budgets, Quest selects query-relevant pages, and KIVI quantizes cached
tensors~\citep{cai2024pyramidkv,tang2024quest,liu2024kivi}.
DuoAttention separates retrieval and streaming
heads~\citep{xiao2025duoattention}. RocketKV combines permanent KV-cache
eviction with subsequent fine-grained sparse
attention~\citep{behnam2025rocketkv}, while ChunkKV retains chunks of
cached tokens to preserve local semantic structure~\citep{liu2025chunkkv}.
These methods reduce the cost of storing or accessing attention KV caches.
\textsc{SparseGDN} skips selected recurrent updates and reduces
intermediate-state traffic during prefill while producing outputs
for every token.

\subsection{Systems Design}

PagedAttention and FlexGen improve serving efficiency through KV-memory
management and tensor placement, respectively~\citep{kwon2023pagedattention,
sheng2023flexgen}. FlashInfer provides a customizable attention engine
for LLM serving~\citep{ye2025flashinfer}. Triton and SparseTIR provide
abstractions for dense tiled and sparse GPU
kernels~\citep{tillet2019triton,ye2023sparsetir}.
\textsc{SparseGDN} couples event selection with deferred update
materialization, compact state storage, and mapped output reads to reduce
both recurrent computation and state traffic.





\end{document}